\documentclass[journal,onecolumn]{IEEEtran}

\usepackage[total={6in,9in}]{geometry}
\usepackage{setspace}

\usepackage{graphicx,psfrag}
\usepackage[cmex10]{amsmath}
\usepackage{amsfonts,amssymb,latexsym,hyperref}
\usepackage[usenames,dvipsnames]{color}
\usepackage{float}
\usepackage[export]{adjustbox} % supplies valign for includegraphics
\usepackage{ifthen} %used for arXiv switch
\newif\ifarxiv
\arxivtrue % set to \arivtrue for long version or \arxivfalse for shorter

\ifarxiv
    \usepackage[nocompress,noadjust]{cite}

    \newcommand{\ocite}[1]{\,\cite{#1}} 
\else
    \usepackage[nospace]{cite}
    \newcommand{\ocite}[1]{} 
\fi

 \renewcommand{\tilde}{\widetilde}
 \renewcommand{\hat}{\widehat}
 
 \newcommand{\Real}{{\mathbb{R}}}
 \newcommand{\Complex}{{\mathbb{C}}}

 \newcommand{\tran}{^{\text{\textsf{T}}}}
 \newcommand{\herm}{^{\text{\textsf{H}}}}
 
 \renewcommand{\vec}[1]{\ensuremath{\boldsymbol{#1}}}
 \newcommand{\hvec}[1]{\ensuremath{\hat{\boldsymbol{#1}}}}
 \newcommand{\uvec}[1]{\ensuremath{\underline{\boldsymbol{#1}}}}
 \newcommand{\defn}{\triangleq}
 
 \newcommand{\Normal}{\mathcal{N}}

 \DeclareMathOperator{\real}{Re}

 \DeclareMathOperator{\E}{E}

 \DeclareMathOperator{\tr}{tr}

 \DeclareMathOperator*{\argmin}{arg\,min}
 
 \DeclareMathOperator{\prox}{prox}
 
\newcommand{\vx}{\vec{x}}
\newcommand{\vy}{\vec{y}}
\newcommand{\vv}{\vec{v}}

\newcommand{\ml}{_{\text{\sf ml}}}
\newcommand{\map}{_{\text{\sf map}}}
\newcommand{\mmse}{_{\text{\sf mmse}}}
\newcommand{\red}{_{\text{\sf red}}}
\newcommand{\pnp}{_{\text{\sf pnp}}}

\newcommand{\proxloss}{\vec{h}}
\newcommand{\proxreg}{\prox_\phi}
\newcommand{\denoiser}{\vec{f}}
\newcommand{\denoisertdt}{\vec{f}_{\text{\sf tdt}}}
\newcommand{\denoisermmse}{\denoiser\mmse}
\newcommand{\denoisertheta}{\denoiser_{\vec{\theta}}}

\newcommand{\px}{p_{\text{\sf x}}}
\newcommand{\pxhat}{\hat{\px}}
\newcommand{\pxsmooth}{\tilde{\px}}

 \newcommand{\textb}[1]{\textcolor{black}{#1}}

 \newcommand{\makeblue}{\color{black}}
 
 \newcommand{\makeblack}{\color{black}}

\begin{document}
\setlength{\arraycolsep}{0.8mm}

\ifarxiv
\title{Plug-and-Play Methods for Magnetic Resonance Imaging (long version)}
\else
\title{Plug-and-Play Methods for Magnetic Resonance Imaging}
\fi

\author{Rizwan Ahmad\thanks{R. Ahmad and S. Liu are with the 
Department of Biomedical Engineering, 
The Ohio State University,
Columbus OH, 43210, USA, 
e-mail: ahmad.46@osu.edu and liu.6221@osu.edu.},
        Charles A. Bouman\thanks{C. Bouman and S. Chan are with the
School of Electrical and Computer Engineering,
Purdue University,
West Lafayette, IN, 47907, USA, 
e-mail: bouman@purdue.edu and stanchan@purdue.edu.},
        Gregery T. Buzzard\thanks{G. Buzzard is with the Department of Mathematics, Purdue University, West Lafayette, IN, 47907, USA, 
e-mail: buzzard@purdue.edu.}, 
        Stanley Chan,\\
        Sizhuo Liu,
        Edward T. Reehorst\thanks{E. Reehorst and P. Schniter are with the 
Department of Electrical and Computer Engineering, 
The Ohio State University,
Columbus OH, 43210, USA, 
e-mail: reehorst.3@osu.edu and schniter.1@osu.edu.},
        and
        Philip Schniter}% <-this % stops a space

\date{}
\maketitle

\ifarxiv
\else
    \doublespacing
\fi

\begin{abstract}
Magnetic Resonance Imaging (MRI) is a non-invasive diagnostic tool that provides excellent soft-tissue contrast without the use of ionizing radiation. 
Compared to other clinical imaging modalities (e.g., CT or ultrasound), however, the data acquisition process for MRI is inherently slow, which motivates undersampling and thus drives the need for accurate, efficient reconstruction methods from undersampled datasets.
%While methods from compressed sensing and machine learning have provided some successes for MRI reconstruction, there remain gaps in our ability to fully leverage a combination of physics-based inversion methods and machine learning methods.   
%
In this article, we describe the use of ``plug-and-play'' (PnP) algorithms for MRI image recovery. 
%which iterate image denoising and forward-model based image recovery.
%PnP algorithms provide a practical approach to compressive MRI because they decouple image modeling from the forward model, which can change significantly among different scans due to variations in the coil sensitivity maps, sampling patterns, and image resolution. 
%Consequently, state-of-the-art image-denoising techniques, such as those based on DNNs, can be directly exploited for compressive MRI image reconstruction through the PnP methodology.  
%PnP methods are also attractive since they can be described as solving a set of equilibrium equations that generalize optimality conditions used in traditional MAP formulations. 
%
We first describe the \textb{linearly approximated} inverse problem encountered in MRI. 
Then we review several PnP methods, where the unifying commonality is to iteratively call a denoising subroutine as one step of a larger optimization-inspired algorithm.
Next, we describe how the result of the PnP method can be interpreted as a solution to an equilibrium equation, allowing convergence analysis from the equilibrium perspective. 
Finally, we present illustrative examples of PnP methods applied to MRI image recovery.
\end{abstract}

%\begin{IEEEkeywords}
%IEEE, IEEEtran, journal, \LaTeX, paper, template.
%\end{IEEEkeywords}

%\IEEEpeerreviewmaketitle

\section{Introduction}
Magnetic Resonance Imaging (MRI) uses radiofrequency waves to non-invasively evaluate the structure, function, and morphology of soft tissues. MRI has become an indispensable imaging tool for diagnosing and evaluating a host of conditions and diseases. Compared to other clinical imaging modalities (e.g., CT or ultrasound), however, MRI suffers from slow data acquisition.
A typical clinical MRI exam consists of multiple scans and can take more than an hour to complete. 
For each scan, the patient may be asked to stay still for several minutes, with slight motion potentially resulting in image artifacts. 
Furthermore, dynamic applications demand collecting a series of images in quick succession. Due to the limited time window in many dynamic applications (e.g., contrast enhanced MR angiography), it is not feasible to collect fully sampled datasets. 
For these reasons, MRI data is often undersampled.
Consequently, computationally efficient methods to recover high-quality images from undersampled MRI data have been actively researched for the last two decades.

The combination of parallel (i.e., multi-coil) imaging and compressive sensing (CS) has been shown to benefit a wide range of MRI applications \cite{lustig2007sparse}\ocite{fessler2019optimization}, including dynamic applications, and has been included in the default image processing \textb{frameworks} offered by several major MRI vendors. 
More recently, learning-based techniques (e.g., \ocite{jin2017deep,zhu2018image}\cite{hyun2018deep,aggarwal2018model,hauptmann2019real,akccakaya2019scan,knoll2019assessment}\ocite{kunisch2013bilevel,hammernik2018learning,lunz2018adversarial,dave2018solving,wen2019transform}) have been shown to outperform CS methods.
Some of these techniques learn the entire end-to-end mapping from undersampled k-space or aliased images to recovered images (e.g.,  \ocite{zhu2018image}\cite{hauptmann2019real}). 
Considering that the forward model in MRI changes from one dataset to the next, such end-to-end methods must be either trained over a large and diverse data corpus or limited to a specific application. 
Other methods train scan-specific convolutional neural networks (CNN) on a fully-sampled region of k-space and then use it to interpolate missing k-space samples~\cite{akccakaya2019scan}.
\textb{These} methods do not require separate training data but demand a fully sampled k-space region. \textb{Due to the large number of unknown CNN parameters, such methods require a fully sampled region that is larger than that typically acquired in parallel imaging,} limiting the acceleration that can be achieved. 
Other supervised learning methods are inspired by classic variational optimization methods and iterate between data-consistency enforcement and a trained CNN, which acts as a regularizer~\cite{aggarwal2018model}\textb{\ocite{hammernik2018learning}}. 
Such methods require a large number of fully sampled, multi-coil k-space datasets, which may be difficult to obtain in many applications. 
Also, since CNN training occurs in the presence of dataset-specific forward models, generalization from training to test scenarios remains an open question \cite{knoll2019assessment}. 
\ifarxiv\textb{Other learning-based methods have been proposed based on bi-level optimization (e.g., \cite{kunisch2013bilevel}), adversarially learned priors (e.g., \cite{lunz2018adversarial}), and autoregressive priors (e.g., \cite{dave2018solving}).} \fi
Consequently, the integration of learning-based methods into physical inverse problems remains a fertile area of research.  
There are many directions for improvement, including recovery fidelity, computational and memory efficiency, robustness, interpretability, and ease-of-use.

This article focuses on ``plug-and-play'' (PnP) algorithms \cite{venkatakrishnan2013plug}, which alternate image denoising with forward-model based signal recovery. 
PnP algorithms facilitate the use of state-of-the-art image models through their manifestations as image denoisers, whether patch-based (e.g., \cite{dabov2007image}) or deep neural network (DNN) based (e.g., \cite{zhang2017beyond}\ocite{chen2017trainable}).
The fact that PnP algorithms decouple image modeling from forward modeling \textb{has advantages} %is advantageous 
in compressive MRI, where the forward model can change significantly among different scans due to variations in the coil sensitivity maps, sampling patterns, and image resolution. 
Furthermore, fully sampled k-space MRI data is not needed for PnP; the image denoiser can be learned from 
%magnitude-only MRI images, or even 
\textb{MRI image patches, or possibly even magnitude-only patches.}
The objective of this article is two-fold: 
i) to review recent advances in plug-and-play methods, and
ii) to discuss their application to compressive MRI image reconstruction.

The remainder of the paper is organized as follows.  
We first detail the inverse problem encountered in MRI reconstruction. 
We then review several PnP methods, where the unifying commonality is to iteratively call a denoising subroutine as one step of a larger optimization-inspired algorithm.
Next, we describe how the result of the PnP method can be interpreted as a solution to an equilibrium equation, allowing convergence analysis from the equilibrium perspective. 
Finally, we present illustrative examples of PnP methods applied to MRI image recovery.  
\ifarxiv
%ADD THIS LINE AFTER THE PAPER IS ACCEPTED
%A shorter version of this paper will appear in \emph{IEEE Signal Processing Magazine}.
\else For an extended version of this paper \textb{that contains additional references and more in-depth discussions on a variety of topics}, see \cite{ahmad2019plug}.\fi

%%%%%%%%%%%%%%%%%%%%%%%%%%%%%%%%%%%%%%%%%%%%%%%%%%%
\section{Image recovery in compressive MRI}  \label{sec:MRI}

In this section, we describe 
the standard linear inverse problem formulation in MRI.
We acknowledge that more sophisticated formulations exist (see, e.g., 
\ifarxiv{\cite{fessler2010model} }\else{the article by Mariya Doneva in this issue }\fi
for a more careful modeling of physics effects).
Briefly, the measurements are samples of the Fourier transform of the image, where the Fourier domain is often referred to as ``k-space.'' 
The transform can be taken across two or three spatial dimensions and includes an additional temporal dimension in dynamic applications. 
Furthermore, measurements are often collected in parallel from $C\geq 1$ receiver coils. 
In dynamic parallel MRI with Cartesian sampling, the time-$t$ k-space measurements from the $i^\text{th}$ coil take the form
\begin{eqnarray}
\vec{y}_i^{(t)} = \vec{P}^{(t)}\vec{F}\vec{S}_i\vec{x}^{(t)} + \vec{w}_i^{(t)}
\label{eq:mri},
\end{eqnarray}
where $\vec{x}^{(t)}\in\Complex^{N}$ is the vectorized 2D or 3D image at discrete time $t$, 
$\vec{S}_i\in\Complex^{N\times N}$ is a diagonal matrix containing the sensitivity map for the $i^\text{th}$ coil, $\vec{F}\in\Complex^{N\times N}$ is the 2D or 3D discrete Fourier transform (DFT),  
the sampling matrix $\vec{P}^{(t)}\in\Real^{M\times N}$ contains $M$ rows of the $N\times N$ identity matrix,
and $\vec{w}_i^{(t)}\in\Complex^{M}$ is additive white Gaussian noise (AWGN). 
Often the sampling pattern changes across frames $t$. 
The MRI literature often refers to $R\triangleq N/M$ as the ``acceleration rate.'' 
The AWGN assumption, which does not hold for the measured parallel MRI data, is commonly enforced using noise pre-whitening filters, which yields the model \eqref{eq:mri} but with diagonal ``virtual'' coil maps $\vec{S}_i$ \cite{hansen2015image}\ocite{buehrer2007array}.
\ifarxiv{Additional justification of the AWGN model can be found in \cite{macovski1996noise}.}\fi

MRI measurements are acquired using a sequence of measurement trajectories through k-space. 
These trajectories can be Cartesian or non-Cartesian in nature. 
Cartesian trajectories are essentially lines through k-space. 
In the Cartesian case, one k-space dimension (i.e., the frequency encoding) is fully sampled, while the other one or two dimensions (i.e., the phase encodings) are undersampled to reduce acquisition time. 
Typically, one line, or ``readout,'' is collected after every RF pulse, and 
the process is repeated several times to collect adequate samples of k-space. 
Non-Cartesian trajectories include radial or spiral curves, which have the effect of distributing the \textb{samples} among all dimensions of k-space. 
Compared to Cartesian sampling, non-Cartesian sampling provides more efficient coverage of k-space and yields an ``incoherent'' forward operator that is more conducive to compressed-sensing reconstruction \ocite{adcock2017news}. 
But Cartesian sampling remains the method of choice in clinical practice, due to its higher tolerance to system imperfections and an extensive record of success. 

Since the sensitivity map, $\vec{S}_i$, is patient-specific and varies with the location of the coil with respect to the imaging plane, both $\vec{S}_i$ and $\vec{x}^{(t)}$ are unknown in practice. 
Although calibration-free methods have been proposed to estimate $\vec{S}_i\vec{x}^{(t)}$
\ifarxiv{(e.g., \cite{shin2014calibrationless}) }\fi
or to jointly estimate $\vec{S}_i$ and $\vec{x}^{(t)}$\ifarxiv{ (e.g., \cite{ying2007joint}), }\else{, }\fi 
it is more common to first estimate $\vec{S}_i$ through a calibration procedure and then treat $\vec{S}_i$ as known in \eqref{eq:mri}.
Stacking $\{\vec{y}_i^{(t)}\}$, $\{\vec{x}^{(t)}\}$, and $\{\vec{w}_i^{(t)}\}$ into vectors $\vec{y}$, $\vec{x}$, and $\vec{w}$, 
and packing $\{\vec{P}^{(t)}\vec{F}\vec{S}_i\}$ into a known block-diagonal matrix $\vec{A}$, we obtain the linear inverse problem of recovering $\vec{x}$ from
\begin{equation}
\vec{y}=\vec{Ax} + \vec{w}, \quad \vec{w}\sim\Normal(\vec{0},\sigma^2\vec{I})
\label{eq:y},
\end{equation}
where $\Normal(\vec{0},\sigma^2\vec{I})$ denotes a circularly symmetric complex-Gaussian random vector.

%%%%%%%%%%%%%%%%%%%%%%%%%%%%%%%%%%%%%%%%%%%%%%%%%%%
\section{Signal Recovery and Denoising} \label{sec:MAP}

The maximum likelihood (ML) estimate of $\vec{x}$ from $\vec{y}$ in \eqref{eq:y} is $\hvec{x}\ml \defn \arg\max_{\vec{x}} p(\vec{y}|\vec{x})$, 
where $p(\vec{y}|\vec{x})$, the probability density of $\vec{y}$ conditioned on $\vec{x}$, is known as the ``likelihood function.''
The ML estimate is often written in the equivalent form $\hvec{x}\ml = \arg\min_{\vec{x}} \{-\ln p(\vec{y}|\vec{x})\}$.
In the case of $\sigma^2$-variance AWGN $\vec{w}$, we have that $-\ln p(\vec{y}|\vec{x}) = \frac{1}{2\sigma^2} \|\vec{y}-\vec{Ax}\|_2^2 + \mathrm{const}$, and so $\hvec{x}\ml = \arg\min_{\vec{x}} \|\vec{y}-\vec{Ax}\|_2^2$, which can be recognized as least-squares estimation.
Although least-squares estimation can give reasonable performance when $\vec{A}$ is tall and well conditioned, this is rarely the case under moderate to high acceleration (i.e., $R>2$).
With acceleration, it is critically important to exploit prior knowledge of signal structure.

The traditional approach to exploiting such prior knowledge is to formulate and solve an optimization problem of the form 
\begin{align}
\hvec{x} = \argmin_{\vec{x}} \left\{ \frac{1}{2\sigma^2} \| \vec{y}-\vec{A}\vec{x}\|_2^2 + \phi(\vec{x}) \right\}
\label{eq:reg_inverse} ,
\end{align}
where the regularization term $\phi(\vec{x})$ encodes prior knowledge of $\vec{x}$.
In fact, $\hvec{x}$ in \eqref{eq:reg_inverse} can be recognized as the maximum a posteriori (MAP) estimate of $\vec{x}$ under the prior density model $p(\vec{x}) \propto \exp(-\phi(\vec{x}))$.
To see why, recall that the MAP estimate maximizes the posterior distribution $p(\vec{x}|\vec{y})$.
That is, $\hvec{x}\map \defn \arg\max_{\vec{x}} p(\vec{x}|\vec{y})
= \arg\min_{\vec{x}} \{- \ln p(\vec{x}|\vec{y})\}$.
Since Bayes' rule implies that $\ln p(\vec{x}|\vec{y}) = \ln p(\vec{y}|\vec{x}) + \ln p(\vec{x}) - \ln p(\vec{y})$, we have 
\begin{align}
\hvec{x}\map = \arg\min_{\vec{x}} \big\{- \ln p(\vec{y}|\vec{x}) - \ln p(\vec{x})\big\} 
\label{eq:MAP} .
\end{align}
Recalling that the first term in \eqref{eq:reg_inverse} \textb{(i.e., the ``loss'' term)} was \textb{observed} to be $-\ln p(\vec{y}|\vec{x})$ (plus a constant) under AWGN noise, the second term in \eqref{eq:reg_inverse} must obey $\phi(\vec{x})=-\ln p(\vec{x}) + \mathrm{const}$.
We will find this MAP interpretation useful in the sequel.

It is not easy to design good regularizers $\phi$ for use in \eqref{eq:reg_inverse}.  
They must not only mimic the negative log signal-prior, but also enable tractable optimization.
One common approach is to use $\phi(\vec{x})=\lambda\|\vec{\Psi x}\|_1$ with $\vec{\Psi}\herm$ a tight frame \textb{(e.g., a wavelet transform)} and $\lambda>0$ a tuning parameter \cite{liu2016projected}.
Such regularizers are convex, and the $\ell_1$ norm rewards sparsity in the transform outputs $\vec{\Psi x}$ \textb{when used with the quadratic loss}.
\ifarxiv
One could go further and use the composite penalty $\phi(\vec{x})=\sum_{l=1}^D \lambda_l \| \vec{\Psi}_l \vec{x} \|_1$.
Due to the richness of data structure in MRI, especially for dynamic applications, utilizing multiple ($D>1$) linear sparsifying transforms has been shown to improve recovery quality \cite{bilen2010compressed}, but tuning multiple regularization weights $\{\lambda_l\}$ is a non-trivial problem \cite{ahmad2015iteratively}.
\fi

Particular insight comes from considering the special case of $\vec{A}=\vec{I}$, where the measurement vector in \eqref{eq:y} reduces to an AWGN-corrupted version of the image $\vec{x}$, i.e.,
\begin{align}
\vec{z} = \vec{x} + \vec{w}, \quad \vec{w}\sim\Normal(\vec{0},\sigma^2\vec{I})
\label{eq:denoise} .
\end{align}
The problem of recovering $\vec{x}$ from noisy $\vec{z}$, known as ``denoising,'' has been intensely researched for decades.
While it is possible to perform denoising by solving a regularized optimization problem of the form \eqref{eq:reg_inverse} with $\vec{A}=\vec{I}$, 
today's state-of-the-art approaches are either algorithmic (e.g., \cite{dabov2007image}) or DNN-based (e.g., \cite{zhang2017beyond}\ocite{chen2017trainable}).
This begs an important question: can these state-of-the-art denoisers be leveraged for MRI signal reconstruction, by exploiting the connections between 
the denoising problem and \eqref{eq:reg_inverse}?
As we shall see, this is precisely what the PnP methods do.

%%%%%%%%%%%%%%%%%%%%%%%%%%%%%%%%%%%%%%%%%%%%%%%%%%%
\section{Plug-and-Play Methods} \label{sec:methods}

In this section, we review several approaches to PnP signal reconstruction.
What these approaches have in common is that they recover $\vec{x}$ from measurements $\vec{y}$ of the form \eqref{eq:y} by iteratively calling a sophisticated denoiser within a larger optimization or inference algorithm.

%---------------------------------------------------------------------
\subsection{Prox-based PnP} \label{sec:PnP}

To start, let us imagine how the optimization in \eqref{eq:reg_inverse} might be solved.  
Through what is known as ``variable splitting,'' we could introduce a new variable, $\vec{v}$, to decouple the regularizer $\phi(\vec{x})$ from the data fidelity term $\frac{1}{2\sigma^2} \|\vec{y} - \vec{A} \vec{x}\|_2^2$.   
The variables $\vec{x}$ and $\vec{v}$ could then be equated using an external constraint, leading to the constrained minimization problem
\begin{eqnarray}
\hat{\vec{x}} = \argmin_{\vec{x}\in \Complex^N} \min_{\vec{v} \in \Complex^N} \left\{ \frac{1}{2\sigma^2} \|\vec{y} - \vec{A} \vec{x}\|_2^2 + \phi(\vec{v})\right\} \quad \mbox{subject to} \quad \vec{x}=\vec{v}.  
\label{eq:split}
\end{eqnarray}
Equation \eqref{eq:split} suggests an algorithmic solution that alternates between separately estimating $\vx$ and estimating $\vv$, with an additional mechanism to asymptotically enforce the constraint $\vx=\vv$.  

The original PnP method \cite{venkatakrishnan2013plug} is based on the alternating direction method of multipliers (ADMM) \cite{boyd2011distributed}.
For ADMM, \eqref{eq:split} is first reformulated as the ``augmented Lagrangian''
\begin{align}
\min_{\vec{x},\vec{v}} \max_{\vec{\lambda}} \left\{
\frac{1}{2\sigma^2}\|\vec{y}-\vec{Ax}\|_2^2 + \phi(\vec{v}) 
    + \real\{\vec{\lambda}\herm(\vec{x}-\vec{v})\}
    + \frac{1}{2\eta}\|\vec{x}-\vec{v}\|^2
\right\}
\label{eq:augmented} ,
\end{align}
where $\vec{\lambda}$ are Lagrange multipliers and $\eta>0$ is a penalty parameter that affects the convergence speed of the algorithm, but not the final solution.
With $\vec{u}\defn \eta\vec{\lambda}$, \eqref{eq:augmented} can be rewritten as
\begin{align}
\min_{\vec{x},\vec{v}} \max_{\vec{u}} \left\{
\frac{1}{2\sigma^2}\|\vec{y}-\vec{Ax}\|_2^2 + \phi(\vec{v}) 
    + \frac{1}{2\eta}\|\vec{x}-\vec{v}+\vec{u}\|^2 
    - \frac{1}{2\eta}\|\vec{u}\|^2
\right\}
\label{eq:augmented2} .
\end{align}
ADMM solves \eqref{eq:augmented2} by alternating the optimization of $\vec{x}$ and $\vec{v}$ with gradient ascent of $\vec{u}$, i.e.,
\begin{subequations} \label{eq:admm} 
\begin{align}
\vec{x}_k &= \proxloss(\vec{v}_{k-1} - \vec{u}_{k-1};\eta)\\
\vec{v}_k &= \proxreg(\vec{x}_{k} + \vec{u}_{k-1};\eta)\\
\vec{u}_k &= \vec{u}_{k-1} + ( \vec{x}_{k} - \vec{v}_{k}), 
\end{align}
\end{subequations}
where $\proxloss(\vec{z};\eta)$ and $\proxreg(\vec{z};\eta)$, known as 
\ifarxiv
``proximal maps'' (see the tutorial \cite{parikh2013proximal})
\else
``proximal maps,'' 
\fi
are defined as
\begin{align}
\proxreg(\vec{z};\eta) 
&\defn \argmin_{\vec{x}} \left\{ \phi(\vec{x}) + \frac{1}{2\eta}\|\vec{x} - \vec{z}\|^2 \right\} 
\label{eq:proxreg}
\\[10pt]
\proxloss(\vec{z};\eta) 
&\defn \argmin_{\vec{x}} \left\{ \frac{1}{2\sigma^2} \|\vec{y} - \vec{Ax}\|^2 + \frac{1}{2\eta}\|\vec{x} - \vec{z} \|^2 \right\} 
= \prox_{\|\vec{y}-\vec{Ax}\|^2/(2\sigma^2)}(\vec{z};\eta) \\
&= \left(\vec{A}\herm\vec{A}+\frac{\sigma^2}{\eta}\vec{I}\right)^{-1}\left(\vec{A}\herm\vec{y}+\frac{\sigma^2}{\eta}\vec{z}\right)
\label{eq:proxloss} .
\end{align}
Under some weak technical constraints, it can be proven \cite{boyd2011distributed} that when $\phi$ is convex, the ADMM iteration \eqref{eq:admm} converges to $\hat{\vec{x}}$, the global minimum of \eqref{eq:reg_inverse} and \eqref{eq:split}.
\ifarxiv{If $\vec{x}_0$ is a fixed point of the recursion \eqref{eq:admm}, then the initialization $\vec{v}_0=\vec{x}_0$ and $\vec{u}_0=\frac{\eta}{\sigma^2}\vec{A}\herm(\vec{y}-\vec{Ax}_0)$ will guarantee that $\vec{x}_k=\vec{x}_0$ for all $k>0$.}\fi

From the discussion in Section~\ref{sec:MAP}, we immediately recognize $\proxreg(\vec{z};\eta)$ in \eqref{eq:proxreg} as the MAP denoiser of $\vec{z}$ under AWGN variance $\eta$ and signal prior $p(\vec{x})\propto \exp(-\phi(\vec{x}))$.
The key idea behind the original PnP work \cite{venkatakrishnan2013plug} was, in the ADMM recursion \eqref{eq:admm}, to ``plug in'' a powerful image denoising algorithm like \textb{``block-matching and 3D filtering'' (BM3D)} \cite{dabov2007image} in place of the proximal denoiser $\proxreg(\vec{z};\eta)$ from \eqref{eq:proxreg}.
If the plug-in denoiser is denoted by ``$\denoiser$,'' then the PnP ADMM algorithm becomes
\begin{subequations} \label{eq:PnP admm}
\begin{align}
\vec{x}_k &= \proxloss(\vec{v}_{k-1} - \vec{u}_{k-1};\eta)\\
\vec{v}_k &= \denoiser(\vec{x}_{k} + \vec{u}_{k-1})\\
\vec{u}_k &= \vec{u}_{k-1} + ( \vec{x}_{k} - \vec{v}_{k}). 
\end{align}
\end{subequations}
A wide variety of empirical results (see, e.g., \cite{venkatakrishnan2013plug,sreehari2016plug,chan2017plug}\ocite{Venkat2016github}) have demonstrated that, when $\denoiser$ is a powerful denoising algorithm like BM3D, the PnP algorithm \eqref{eq:PnP admm} produces far better recoveries than the regularization-based approach \eqref{eq:admm}.
\ifarxiv{For parallel MRI, the advantages of PnP ADMM were demonstrated in \cite{pouryazdanpanah2019deep}.}\fi
Although the value of $\eta$ does not change the fixed point of the standard ADMM algorithm \eqref{eq:admm}, it affects the fixed point of the PnP ADMM algorithm \eqref{eq:PnP admm} through the ratio $\sigma^2/\eta$ in \eqref{eq:proxloss}.

The success of PnP methods raises important theoretical questions. 
Since $\denoiser$ is not in general the proximal map of any regularizer $\phi$, 
the iterations \eqref{eq:PnP admm} may not minimize a cost function of the form in \eqref{eq:reg_inverse}, and 
\eqref{eq:PnP admm} may not be an implementation of ADMM.
It is then unclear if the iterations \eqref{eq:PnP admm} will converge. 
And if they do converge, it is unclear what they converge to.
The consensus equilibrium framework, which we discuss in Section~\ref{sec:CE}, \textb{aims to provide} answers to these questions. 

The use of a generic denoiser in place of a proximal denoiser can be translated to non-ADMM algorithms, such as FISTA \ocite{beck2009fast}, primal-dual \textb{splitting (PDS)} \ocite{esser2010general}, and others, as in \cite{kamilov2017plug,ono2017primal,sun2019online}. 
Instead of optimizing $\vec{x}$ as in \eqref{eq:PnP admm}, PnP FISTA \cite{kamilov2017plug} uses the iterative update 
\begin{subequations} \label{eq:PnP fista}
\begin{align}
    \vec{z}_k &= \vec{s}_{k-1} - \frac{\eta}{\sigma^2} \vec{A}\herm(\vec{A} \vec{s}_{k-1} - \vy)
        \label{eq:PnP fista 1}\\
    \vec{x}_k &= \denoiser(\vec{z}_k)
        \label{eq:PnP fista 2}\\
    \vec{s}_{k} &= \vec{x}_{k} + \frac{q_{k-1}-1}{q_k}(\vec{x}_k - \vec{x}_{k-1})
        \label{eq:PnP fista 3} ,
\end{align}
\end{subequations}
where 
\eqref{eq:PnP fista 1} is a gradient descent (GD) step on the negative log-likelihood $\frac{1}{2\sigma^2}\|\vec{y}-\vec{Ax}\|^2$ at $\vec{x}=\vec{s}_{k-1}$ with step-size $\eta \textb{\in (0,\sigma^2\|\vec{A}\|_2^{-2})}$, 
\eqref{eq:PnP fista 2} is the plug-in replacement of the usual proximal denoising step in FISTA, and 
\eqref{eq:PnP fista 3} is an acceleration step, where it is typical to use
$q_k = (1+\sqrt{1+4q_{k-1}^2})/2$ and $q_0=1$.
\ifarxiv{If $\vec{x}_0$ is a fixed point of the recursion \eqref{eq:PnP fista}, then the initialization $\vec{s}_0=\vec{x}_0$ will guarantee that $\vec{x}_k=\vec{x}_0$ for all $k>0$.}\fi

\ifarxiv
\makeblue
PnP PDS \cite{ono2017primal} uses the iterative update
\begin{subequations} \label{eq:PnP PDS}
\begin{align}
    \vec{x}_k &= \vec{f}\Big(\vec{x}_{k-1} - \frac{\eta}{\sigma^2}\vec{A}\herm\vec{v}_{k-1}\Big) 
        \label{eq:PnP PDS 1}\\
    \vec{v}_k &= \gamma \vec{v}_{k-1} + (1-\gamma) \Big( \vec{A}(2\vec{x}_k-\vec{x}_{k-1})- \vec{y}\Big) 
        \label{eq:PnP PDS 2}
\end{align}
\end{subequations}
where $\eta>0$ is a stepsize, 
$\gamma\in(0,1)$ is a relaxation parameter chosen such that
$\gamma\leq \eta/(\eta+\sigma^2\|\vec{A}\|_2^{-2})$,
and $\vec{f}(\vec{z})$ in \eqref{eq:PnP PDS 1} is the plug-in replacement of the usual proximal denoiser $\proxreg(\vec{z};\eta)$.
If $\vec{x}_0$ is a fixed point of the recursion \eqref{eq:PnP PDS}, then the initialization $\vec{v}_0=\vec{Ax}_0-\vec{y}$ will guarantee that $\vec{x}_k=\vec{x}_0$ for all $k>0$.\makeblack
\fi

Comparing  PnP ADMM \eqref{eq:PnP admm} to PnP FISTA \eqref{eq:PnP fista}\ifarxiv\textb{~and PnP PDS} \eqref{eq:PnP PDS}\fi, one can see that they differ in how the data fidelity term $\frac{1}{2\sigma^2}\|\vec{y}-\vec{Ax}\|^2$ is handled:
PnP ADMM uses the proximal update \eqref{eq:proxloss}, 
while PnP FISTA and PnP PDS use the GD step \eqref{eq:PnP fista 1}.
%If the proximal update is solved exactly, then PnP ADMM usually requires many fewer iterations than PnP FISTA.
%But exactly 
In most cases, solving the proximal update \eqref{eq:proxloss} is much more computationally costly than taking a GD step \eqref{eq:PnP fista 1}.
%Another option is to 
\textb{Thus, with ADMM, it is common to} approximate the proximal update \eqref{eq:proxloss} using, e.g., 
several iterations of the conjugate gradient (CG) algorithm or GD, which should reduce the per-iteration complexity 
of \eqref{eq:PnP admm} but may increase the number of iterations.
%In the end, the most efficient approach will depend on the relative complexity of the denoiser $\denoiser$ and the (exact or approximated) proximal update \eqref{eq:proxloss}.
\textb{But even with these approximations of \eqref{eq:proxloss}, PnP ADMM is usually close to ``convergence'' after 10-50 iterations (e.g., see Figure~\ref{fig:nmse_iter}).}

\textb{An important difference between the aforementioned flavors of PnP is that the stepsize $\eta$ is constrained in FISTA but not in ADMM or PDS. 
Thus, PnP FISTA restricts the range of reachable fixed points relative to PnP ADMM and PnP PDS.}

%---------------------------------------------------------------------
\subsection{The balanced FISTA approach} \label{sec:bFISTA}

In Section~\ref{sec:MAP}, when discussing the optimization problem \eqref{eq:reg_inverse}, the regularizer $\phi(\vec{x})=\lambda\|\vec{\Psi x}\|_1$ was mentioned as a popular option,
where $\vec{\Psi}$ is often a wavelet transform.
The resulting optimization problem,
\begin{align}
\hvec{x}
= \argmin_{\vec{x}} \left\{ \frac{1}{2\sigma^2} \|\vec{y}-\vec{Ax}\|_2^2 
    + \lambda \|\vec{\Psi x}\|_1 \right\}
\label{eq:analysis} ,
\end{align}
is said to be stated in ``analysis'' form
\ifarxiv\cite{fessler2019optimization}\else{\textb{(see \cite{fessler2020optimization} in this special issue)}}\fi. The proximal denoiser associated with \eqref{eq:analysis} has the form
\begin{align}
\proxreg(\vec{z};\eta)=\argmin_{\vec{x}} \left\{ \lambda\|\vec{\Psi x}\|_1+ \frac{1}{2\eta}\|\vec{x}-\vec{z}\|^2\right\}
\label{eq:proxanal}.
\end{align}
When $\vec{\Psi}$ is orthogonal, it is well known that
$\proxreg(\vec{z};\eta)=\denoisertdt(\vec{z};\lambda\eta)$,
where
\begin{align}
\denoisertdt(\vec{z};\tau)
\defn \vec{\Psi}\herm\textrm{soft-thresh}(\vec{\Psi z};\tau) 
\label{eq:denoisertdt}  
\end{align}
is the ``transform-domain thresholding'' denoiser 
with $[\textrm{soft-thresh}(\vec{u},\tau)]_n\defn\max\big\{0,\frac{|u_n|-\tau}{|u_n|}\big\}u_n$.
The denoiser \eqref{eq:denoisertdt} is very efficient to implement, since it amounts to little more than computing forward and reverse transforms.

In practice, \eqref{eq:analysis} yields much better results with \emph{non}-orthogonal $\vec{\Psi}$, such as when $\vec{\Psi}\herm$ is a tight frame (see, e.g., the references in \cite{ting2017fast}).  
In the latter case, $\vec{\Psi}\herm\vec{\Psi}=\vec{I}$ with tall $\vec{\Psi}$.
But, for general tight frames $\vec{\Psi}\herm$, the proximal denoiser \eqref{eq:proxanal} has no closed-form solution.
What if we simply plugged the transform-domain thresholding denoiser \eqref{eq:denoisertdt} into an algorithm like ADMM or FISTA? 
How can we interpret the resulting approach?
Interestingly, as we describe below, if \eqref{eq:denoisertdt} is used in PnP FISTA, then it does solve a convex optimization problem, although one with a different form than \eqref{eq:reg_inverse}.
This approach was independently proposed in \cite{liu2016projected} and \cite{ting2017fast}, where in the latter it was referred to as ``balanced FISTA'' (bFISTA) and applied to parallel cardiac MRI.
Notably, bFISTA was proposed before the advent of PnP FISTA.
More details are provided below.

The optimization problem \eqref{eq:analysis} can be stated in constrained ``synthesis'' form as
\begin{align}
\hvec{x}
= \vec{\Psi}\herm\hvec{\alpha}
\quad
\text{for}
\quad
\hvec{\alpha}
= \argmin_{\vec{\alpha}\in\mathrm{range}(\vec{\Psi})} \left\{ \frac{1}{2\sigma^2} \|\vec{y}-\vec{A\Psi}\herm\vec{\alpha}\|_2^2 
    + \lambda \|\vec{\alpha}\|_1 \right\}
\label{eq:synthesis} ,
\end{align}
where $\vec{\alpha}$ are transform coefficients.
Introducing a new parameter $\beta$, the problem \eqref{eq:synthesis} can be extended to a 1-parameter family of unconstrained problems
%Then, as $\beta\rightarrow\infty$ below, \eqref{eq:synthesis} can be expressed in the unconstrained form 
\begin{align}
\hvec{x}
= \vec{\Psi}\herm\hvec{\alpha}
\quad
\text{for}
\quad
\hvec{\alpha}
= \argmin_{\vec{\alpha}} \left\{ \frac{1}{2\sigma^2} \|\vec{y}-\vec{A\Psi}\herm\vec{\alpha}\|_2^2
    + \frac{\beta}{2} \|\vec{P}_{\vec{\Psi}}^\perp\vec{\alpha}\|_2^2
    + \lambda \|\vec{\alpha}\|_1 \right\}
\label{eq:balanced} 
\end{align}
with projection matrix $\vec{P}_{\vec{\Psi}}^\perp \defn \vec{I}-\vec{\Psi\Psi}\herm$,
and with \eqref{eq:synthesis} corresponding to the limiting case of \eqref{eq:balanced} with $\beta\rightarrow\infty$.
In practice, it is not possible to take $\beta\rightarrow\infty$ and, for finite values of $\beta$, the problems \eqref{eq:synthesis} and \eqref{eq:balanced} are not equivalent.
However, problem \eqref{eq:balanced} under finite $\beta$ is interesting to consider in its own right, and it is sometimes referred to as the ``balanced'' approach \ocite{shen2011accelerated}.
If we solve \eqref{eq:balanced} using FISTA with step-size $\eta>0$ (recall \eqref{eq:PnP fista 1}) and choose the particular value $\beta=1/\eta$ then, remarkably, the resulting algorithm takes the form of PnP FISTA \eqref{eq:PnP fista} with $\denoiser(\vec{z})=\denoisertdt(\vec{z};\lambda)$.
This particular choice of $\beta$ is motivated by computational efficiency (since it leads to the use of $\denoisertdt$) rather than recovery performance.
Still, as we demonstrate in Section~\ref{sec:sim}, it performs relatively well.

%---------------------------------------------------------------------
\subsection{Regularization by denoising} \label{sec:RED}

Another PnP approach, proposed by Romano, Elad, and Milanfar in \cite{romano2017little},  recovers $\vec{x}$ from measurements $\vec{y}$ in \eqref{eq:y} by finding the $\hvec{x}$ that solves the optimality condition\footnote{We begin our discussion of RED by focusing on the real-valued case, \textb{as in \cite{romano2017little} and \cite{reehorst2019regularization}}, but later we extend the RED algorithms to the complex-valued case of interest in MRI.}
\begin{equation}
\vec{0} 
= \frac{1}{\sigma^2}\vec{A}\tran(\vec{A}\hvec{x}-\vec{y}) + \frac{1}{\eta}(\hvec{x}-\denoiser(\hvec{x}))
\label{eq:fpRED} ,
\end{equation}
where $\denoiser$ is an arbitrary (i.e., ``plug in'') image denoiser and $\eta>0$ is a tuning parameter.
%\ifarxiv\footnote{\textb{If the images live on a manifold of dimension $D\ll N$, then $T\eta^{D/2}\propto 1$}}\fi
In \cite{romano2017little}, several algorithms were proposed to solve \eqref{eq:fpRED}.
Numerical experiments in \cite{romano2017little} suggest that, when $\denoiser$ is a sophisticated denoiser (like BM3D) and $\eta$ is well tuned, the solutions $\hvec{x}$ to \eqref{eq:fpRED} are state-of-the-art, similar to those of PnP ADMM.

The approach \eqref{eq:fpRED} was coined ``regularization by denoising'' (RED) in \cite{romano2017little} because,  under certain conditions, the $\hvec{x}$ that solve \eqref{eq:fpRED} are the solutions to the regularized least-squares problem
\begin{align}
\hvec{x} = \argmin_{\vec{x}} \left\{ \frac{1}{2\sigma^2} \|\vec{y}-\vec{Ax}\|^2 + \phi\red(\vec{x}) \right\}
\text{~with~} \phi\red(\vec{x}) \defn \frac{1}{2\eta} \vec{x}\tran\big(\vec{x}-\denoiser(\vec{x})\big) 
\label{eq:optRED} ,
\end{align}
where the regularizer $\phi\red$ is explicitly constructed from the plug-in denoiser $\denoiser$.
But what are these conditions?
Assuming that $\denoiser$ is differentiable almost everywhere,
it was shown in \cite{reehorst2019regularization} that the solutions of \eqref{eq:fpRED} correspond to those of \eqref{eq:optRED} when 
i)  $\denoiser$ is locally homogeneous\footnote{Locally homogeneous means that $(1+\epsilon)\denoiser(\vec{x}) = \denoiser\big((1+\epsilon)\vec{x}\big)$ for all $\vec{x}$ and sufficiently small nonzero $\epsilon$.} and
ii) $\denoiser$ has a symmetric Jacobian matrix (i.e., $[J\denoiser(\vec{x})]\tran = J\denoiser(\vec{x})~\forall \vec{x}$).
But it was demonstrated in \cite{reehorst2019regularization} that these properties are \emph{not} satisfied by popular image denoisers,
such as the median filter, transform-domain thresholding, NLM, BM3D, TNRD, and DnCNN.
Furthermore, it was proven in \cite{reehorst2019regularization} that if the Jacobian of $\denoiser$ is non-symmetric, then there does not exist \emph{any} regularizer $\phi$ under which the solutions of \eqref{eq:fpRED} minimize a regularized loss of the form in \eqref{eq:reg_inverse}.

One may then wonder how to justify \eqref{eq:fpRED}.
In \cite{reehorst2019regularization}, Reehorst and Schniter proposed an explanation for \eqref{eq:fpRED} based on 
``score matching'' \ocite{hyvarinen2005estimation}, which we now summarize.
Suppose we are given a large corpus of training images $\{\vec{x}_t\}_{t=1}^T$, from which we could build the empirical prior model
\begin{align*}
\pxhat(\vec{x}) 
&\defn \frac{1}{T}\sum_{t=1}^T \delta(\vec{x}-\vec{x}_t) ,
\end{align*}
where $\delta$ denotes the Dirac delta.
Since images are known to exist outside $\{\vec{x}_t\}_{t=1}^T$, it is possible to build an improved prior model $\pxsmooth$ using kernel density estimation (KDE), i.e.,
\begin{align}
\pxsmooth(\vec{x};\eta) 
&\defn \frac{1}{T}\sum_{t=1}^T \Normal(\vec{x};\vec{x}_t,\eta\vec{I}) ,
\label{eq:KDE}
\end{align}
where $\eta>0$ is a tuning parameter.
If we adopt $\pxsmooth$ as the prior model for $\vec{x}$, then the MAP estimate of $\vec{x}$ (recall \eqref{eq:MAP}) becomes
\begin{align}
\hvec{x} = \arg\min_{\vec{x}} \left\{\frac{1}{2\sigma^2}\|\vec{y}-\vec{Ax}\|^2 - \ln \pxsmooth(\vec{x};\eta)\right\} 
\label{eq:MAPkde} .
\end{align}
Because $\ln \pxsmooth$ is differentiable, the solutions to \eqref{eq:MAPkde} must obey
\begin{align}
\vec{0} = \frac{1}{\sigma^2}\vec{A}\tran(\vec{A}\hvec{x}-\vec{y}) - \nabla \ln \pxsmooth(\hvec{x};\eta) 
\label{eq:fpMAPkde}.
\end{align}
A classical result known as ``Tweedie's formula'' \ocite{robbins1956empirical,efron2011tweedie} says that 
\begin{align}
\nabla \ln \pxsmooth(\vec{z};\eta) 
&= \frac{1}{\eta}\big(\denoisermmse(\vec{z};\eta)-\vec{z}\big) 
\label{eq:Tweedie},
\end{align}
where $\denoisermmse(\cdot;\eta)$ is the minimum mean-squared error (MMSE) denoiser under the prior $\vec{x}\sim\pxhat$ and $\eta$-variance AWGN.
That is, $\denoisermmse(\vec{z})=\E\{\vec{x}\,|\, \vec{z}\}$, where $\vec{z}=\vec{x}+\Normal(\vec{0},\eta\vec{I})$ and $\vec{x}\sim\pxhat$. 
Applying \eqref{eq:Tweedie} to \eqref{eq:fpMAPkde}, the MAP estimate $\hvec{x}$ under the KDE prior $\pxsmooth$ obeys
\begin{align}
\vec{0} = \frac{1}{\sigma^2}\vec{A}\tran(\vec{A}\hvec{x}-\vec{y}) + \frac{1}{\eta}\big(\hvec{x}-\denoisermmse(\hvec{x};\eta)\big) 
\label{eq:fpREDmmse},
\end{align}
which matches the RED condition \eqref{eq:fpRED} when $\denoiser=\denoisermmse(\cdot;\eta)$.
Thus, if we could implement the MMSE denoiser $\denoisermmse$ for a given training corpus $\{\vec{x}_t\}_{t=1}^T$, then RED provides a way to 
compute the MAP estimate of $\vec{x}$ under the KDE prior $\pxsmooth$.

Although the MMSE denoiser $\denoisermmse$ can be expressed in closed form (see \cite[eqn.~(67)]{reehorst2019regularization}), it is not practical to implement for large $T$.
Thus the question remains: Can the RED approach \eqref{eq:fpRED} also be justified for non-MMSE denoisers $\denoiser$, 
especially those that are not locally homogeneous or Jacobian-symmetric?
As shown in \cite{reehorst2019regularization}, the answer is yes.
Consider a practical denoiser $\denoisertheta$ parameterized by tunable weights $\vec{\theta}$ (e.g., a DNN). 
A typical strategy is to choose $\vec{\theta}$ to minimize the mean-squared error on $\{\vec{x}_t\}_{t=1}^T$, i.e., set
$\hvec{\theta} 
= \argmin_{\vec{\theta}} \E\{\|\vec{x}-\denoisertheta(\vec{z})\|^2\}$,
where the expectation is taken over $\vec{x}\sim\pxhat$ and $\vec{z}=\vec{x}+\Normal(\vec{0},\eta\vec{I})$.
By the MMSE orthogonality principle, we have
\begin{align}
\E\big\{\|\vec{x}-\denoisertheta(\vec{z})\|^2\big\} 
= \E\big\{\|\vec{x}-\denoisermmse(\vec{z};\eta)\|^2\big\} + \E\big\{\|\denoisermmse(\vec{z};\eta)-\denoisertheta(\vec{z})\|^2\big\} ,
\end{align}
and so we can write
\begin{align}
\hvec{\theta} 
&= \argmin_{\vec{\theta}} \E\big\{\|\denoisermmse(\vec{z};\eta)-\denoisertheta(\vec{z})\|^2\big\} \\
&= \argmin_{\vec{\theta}} \E\left\{\left\|\nabla \ln \pxsmooth(\vec{z};\eta)-\frac{1}{\eta}\big(\denoisertheta(\vec{z})-\vec{z}\big)\right\|^2\right\} 
\label{eq:scorematch} ,
\end{align}
where \eqref{eq:scorematch} follows from \eqref{eq:Tweedie}.
Equation \eqref{eq:scorematch} says that choosing $\vec{\theta}$ to minimize the MSE is equivalent to choosing $\vec{\theta}$ so that $\frac{1}{\eta}(\denoisertheta(\vec{z})-\vec{z})$ best matches the ``score'' $\nabla \ln \pxsmooth(\vec{z};\eta)$. 
\ifarxiv
This is an instance of the ``score matching'' framework, as described in \cite{hyvarinen2005estimation}.
\fi

In summary, the RED approach \eqref{eq:fpRED} approximates the KDE-MAP approach \eqref{eq:fpMAPkde}-\eqref{eq:fpREDmmse} by using a plug-in denoiser $\denoiser$ to approximate the MMSE denoiser $\denoisermmse$.
When $\denoiser=\denoisermmse$, RED exactly implements MAP-KDE, but with a practical $\denoiser$, RED implements a score-matching approximation of MAP-KDE.
Thus, a more appropriate title for RED might be ``score matching by denoising.''

Comparing the RED approach from this section to the prox-based PnP approach from Section~\ref{sec:PnP}, we see that 
RED starts with the KDE-based MAP estimation problem \eqref{eq:MAPkde} and replaces the $\pxhat$-based MMSE denoiser $\denoisermmse$ with a plug-in denoiser $\denoiser$,
while PnP ADMM starts with the $\phi$-based MAP estimation problem \eqref{eq:reg_inverse} and replaces the $\phi$-based MAP denoiser $\proxreg$ from \eqref{eq:proxreg} with a plug-in denoiser $\denoiser$.
It has recently been demonstrated that, when the prior is constructed from image examples, MAP recovery often leads to sharper, more natural looking image recoveries than MMSE recovery \cite{sonderby2016amortised}\ocite{chang2017one,meinhardt2017learning,bigdeli2019image}.
Thus it is interesting that RED offers an approach to MAP-based recovery using MMSE denoising, which is much easier to implement than MAP denoising \cite{sonderby2016amortised}\ocite{chang2017one,meinhardt2017learning,bigdeli2019image}.

Further insight into the difference between RED and prox-based PnP can be obtained by considering the case of symmetric linear denoisers, i.e., $\denoiser(\vec{z})=\vec{Wz}$ with $\vec{W}=\vec{W}\tran$, where we will also assume that $\vec{W}$ is invertible.
Although such denoisers are far from state-of-the-art, they are useful for interpretation.
It is easy to show \cite{chan2019performance} that $\denoiser(\vec{z})=\vec{Wz}$ is the proximal map of 
$\phi(\vec{x}) = \frac{1}{2\eta}\vec{x}\tran(\vec{W}^{-1}-\vec{I})\vec{x}$,
i.e., that $\proxreg(\vec{z};\eta)=\vec{Wz}$, recalling \eqref{eq:proxreg}.
With this proximal denoiser, we know that the prox-based PnP algorithm solves the optimization problem
\begin{align}
\hvec{x}\pnp = \textb{\argmin}_{\vec{x}} \left\{\frac{1}{2\sigma^2}\|\vec{y}-\vec{Ax}\|^2 
            + \frac{1}{2\eta}\vec{x}\tran(\vec{W}^{-1}-\vec{I})\vec{x}\right\} 
\label{eq:PnP lin}.
\end{align}
Meanwhile, since $\denoiser(\vec{z})=\vec{Wz}$ is both locally homogeneous and Jacobian-symmetric, we know from \eqref{eq:optRED} that the RED under this $\denoiser$ solves the optimization problem
\begin{align}
\hvec{x}\red = \textb{\argmin}_{\vec{x}} \left\{\frac{1}{2\sigma^2}\|\vec{y}-\vec{Ax}\|^2 
            + \frac{1}{2\eta}\vec{x}\tran(\vec{I}-\vec{W})\vec{x}\right\} 
\label{eq:red lin}.
\end{align}
By comparing \eqref{eq:PnP lin} and \eqref{eq:red lin}, we see a clear difference between RED and prox-based PnP.
\ifarxiv %%%%%%%%%%%%%%%%%%%%%%%%%%%%%%%%%%%%%%%%%%%%%%%%%%%%%%%%%%%%%%
Using $\vec{u}\defn\vec{W}^{1/2}\vec{x}$, it can be seen that we can rewrite \eqref{eq:red lin} as
\begin{align}
\hvec{x}\red &= \vec{W}^{-1/2}\hvec{u} \text{~~for~~}
\hvec{u} = \textb{\argmin}_{\vec{u}} \left\{\frac{1}{2\sigma^2}\|\vec{y}-\vec{AW}^{-1/2}\vec{u}\|^2 
            + \frac{1}{2\eta}\vec{u}\tran(\vec{W}^{-1}-\vec{I})\vec{u}\right\} 
\label{eq:red lin2},
\end{align}
which has the same regularizer as \eqref{eq:PnP lin} but a different loss and additional post-processing.
\fi %%%%%%%%%%%%%%%%%%%%%%%%%%%%%%%%%%%%%%%%%%%%%%%%%%%%%%%%%%%%%%%%%%%
Section~\ref{sec:CE RED} compares RED to prox-based PnP from yet another perspective: consensus equilibrium.

So far, we have described RED as solving for $\hvec{x}$ in \eqref{eq:fpRED}.
But how exactly is this accomplished?
In the original RED paper \cite{romano2017little}, three algorithms were proposed to solve \eqref{eq:fpRED}: 
GD, inexact ADMM, and a ``fixed point'' heuristic that was later recognized \cite{reehorst2019regularization} as a special case of the proximal gradient (PG) algorithm \ocite{beck2009gradient,combettes2011proximal}.
Generalizations of PG RED were proposed in \cite{reehorst2019regularization}.
The fastest among them is the accelerated-PG RED algorithm, which uses the iterative update\footnote{Although the development of RED up to this point has assumed real-valued quantities for simplicity, the RED equations \eqref{eq:red apg} and \eqref{eq:red gd 1} may be used with complex-valued quantities.}
\begin{subequations} \label{eq:red apg}
\begin{align}
\vec{x}_k &= \proxloss(\vec{v}_{k-1};\eta/L) 
\label{eq:red apg 1}\\
\vec{z}_k &= \vec{x}_k + \frac{q_{k-1}-1}{q_k} (\vec{x}_k-\vec{x}_{k-1}) 
\label{eq:red apg 2}\\
\vec{v}_k &= \frac{1}{L}\denoiser(\vec{z}_k) + \bigg(1-\frac{1}{L}\bigg)\vec{z}_k
\label{eq:red apg 3} ,
\end{align}
\end{subequations}
where $\proxloss$ was defined in \eqref{eq:proxloss}, 
line \eqref{eq:red apg 2} uses the same acceleration as PnP FISTA \eqref{eq:PnP fista 2},
and $L>0$ is a design parameter that can be related to the Lipschitz constant of $\phi\red(\cdot)$ from \eqref{eq:optRED} (see \cite[Sec.V-C]{reehorst2019regularization}).
When $L=1$ and $q_k=1~\forall k$, \eqref{eq:red apg} reduces to the ``fixed point'' heuristic from \cite{romano2017little}.
To reduce the implementation complexity of $\proxloss$, one could replace \eqref{eq:red apg 1} with the 
GD step
\begin{align}
\vec{x}_k &= \vec{x}_{k-1} - \frac{\eta}{L \sigma^2} \vec{A}\herm(\vec{Av}_{k-1}-\vec{y})
\label{eq:red gd 1} ,
\end{align}
which achieves a similar complexity reduction as when going from PnP ADMM to PnP FISTA (as discussed in Section~\ref{sec:PnP}).
The result would be an ``accelerated GD'' \ocite{nesterov1983method} form of RED.
Convergence of the RED algorithms will be discussed in Section~\ref{sec:CE RED}.

%---------------------------------------------------------------------
\ifarxiv
\subsection{Approximate Message Passing}

The approximate message passing (AMP) algorithm \ocite{donoho2009message} estimates the signal $\vec{x}$ using the iteration
\begin{subequations}\label{eq:amp}
\begin{align}
\vec{v}_k &= \frac{\sqrt{N}}{\|\vec{A}\|_F} (\vec{y}-\vec{Ax}_{k-1}) 
        + \frac{1}{M}\tr\{J \vec{f}_{k-1}(\vec{z}_{k-1})\}\vec{v}_{k-1} 
        \label{eq:amp1}\\
\vec{z}_k &= \vec{x}_{k-1} + \vec{A}\herm \vec{v}_k 
        \label{eq:amp2}\\
\vec{x}_k &= \vec{f}_k(\vec{z}_k) 
\end{align}
\end{subequations}
starting with $\vec{v}_0=\vec{0}$ and some $\vec{x}_0$.
In \eqref{eq:amp}, $M$ and $N$ refer to the dimensions of $\vec{A}\in\Real^{M\times N}$, $J$ denotes the Jacobian operator,
and the denoiser $\vec{f}_k$ may change with iteration $k$.
In practice, the trace of the Jacobian in \eqref{eq:amp1} can be approximated using \ocite{ramani2008monte}
\begin{align}
\tr\{J \vec{f}_k(\vec{z}_k)\}
\approx \vec{p}\herm\big[\vec{f}_k(\vec{z}_k+\epsilon\vec{p})-\vec{f}_k(\vec{z}_k)\big]/\epsilon % COMPLEX CASE
\end{align}
using small $\epsilon>0$ and random $\vec{p}\sim\Normal(\vec{0},\vec{I})$. 
The original AMP paper \ocite{donoho2009message} considered only real-valued quantities and proximal denoising 
(i.e., $\vec{f}_k(\vec{z})=\proxreg(\vec{z};\eta_k)$ for some $\eta_k>0$);
the complex-valued case was first described in \cite{schniter2010turbo} and
the PnP version of AMP shown in \eqref{eq:amp} was proposed in \ocite{metzler2015bm3d}\cite{metzler2016from} under the name ``denoising AMP.''
This approach was applied to MRI in \cite{eksioglu2018denoising}, but the results are difficult to reproduce.

The rightmost term in \eqref{eq:amp1} is known as the ``Onsager correction term,'' with roots in statistical physics \ocite{zdeborova2016statistical}.
If we remove this term from \eqref{eq:amp1}, the resulting recursion is no more than unaccelerated FISTA (also known as ISTA \ocite{chambolle1998nonlinear}), i.e., \eqref{eq:PnP fista} with $q_k=1~\forall k$ and a particular stepsize $\eta$.
The Onsager term is what gives AMP the special properties discussed below.

When $\vec{A}$ is i.i.d.\ Gaussian and the denoiser $\vec{f}_k$ is Lipschitz, the PnP AMP iteration \eqref{eq:amp} has remarkable properties in the large-system limit (i.e., $M,N\rightarrow\infty$ with fixed $M/N$), as proven in \cite{berthier2019state} for the real-valued case. 
First,
\begin{align}
\vec{z}_k &= \vec{x}^0 + \vec{e} \text{~~for~~} \vec{e}\sim\Normal(\vec{0},\eta_k\vec{I}) \text{~~with~~} \eta_k = \lim_{M\rightarrow\infty}\|\vec{v}_k\|^2/M 
\label{eq:zk}.
\end{align}
That is, the denoiser input $\vec{z}_k$ behaves like an AWGN corrupted version of the true image $\vec{x}^0$.
Moreover, as $N\rightarrow\infty$, the mean-squared of AMP's estimate $\vec{x}_k$ is exactly predicted by the ``state evolution''
\begin{subequations}\label{eq:se}
\begin{align}
\mathcal{E}(\eta_k) &= \lim_{N\rightarrow\infty}\frac{1}{N} \E\left\{\left\|\vec{f}_k(\vec{x}^0+\Normal(\vec{0},\eta_k\vec{I}))-\vec{x}^0\right\|^2\right\} \\
\eta_{k+1} &= \sigma^2 + \frac{N}{M}\mathcal{E}(\eta_k) .
\end{align}
\end{subequations}
Furthermore, if $\vec{f}_k$ is the MMSE denoiser  
(i.e., $\vec{f}_k(\vec{z})=\E\{\vec{x}\,|\,\vec{z}=\vec{x}^0+\Normal(\vec{0},\eta_k\vec{I})\}$)
and the state-evolution \eqref{eq:se} has a unique fixed-point, 
then $\mathcal{E}(\eta_k)$ converges to the minimum MSE. 
Or, if $\vec{f}_k$ is the proximal denoiser $\vec{f}_k(\vec{z})=\proxreg(\vec{z};\eta_k)$ 
and the state-evolution \eqref{eq:se} has a unique fixed-point, 
then AMP's $\vec{x}_k$ converges to the MAP solution, even if $\phi$ is non-convex.

The remarkable properties of AMP are only guaranteed to manifest with large i.i.d.\ Gaussian $\vec{A}$.
For this reason, a ``vector AMP'' (VAMP) algorithm was proposed \ocite{rangan2016vector} with a rigorous state-evolution and denoiser-input property \eqref{eq:zk} that hold under the larger class of right-rotationally invariant (RRI)\footnote{Matrix $\vec{A}$ is said to be RRI if its singular-value decomposition $\vec{A}=\vec{USV}\herm$ has Haar $\vec{V}$, i.e., $\vec{V}$ distributed uniformly over the group of orthogonal matrices for any $\vec{U}$ and $\vec{S}$.  
Note that i.i.d.\ Gaussian is a special case of RRI where $\vec{U}$ is also Haar and the singular values in $\vec{S}$ have a particular distribution.}
matrices $\vec{A}$.
Later, a PnP version of VAMP was proposed \ocite{schniter2017denoising} and rigorously analyzed under Lipschitz $\vec{f}_k$ \ocite{fletcher2018plug}.
VAMP can be understood as the symmetric variant \ocite{he2016convergence} of ADMM with adaptive penalty $\eta$ \ocite{fletcher2016expectation}. 

Both AMP and VAMP assume that the matrix $\vec{A}$ is a typical realization of a random matrix drawn \emph{independently} 
of the signal $\vec{x}$, so that---with high probability---multiplication-by-$\vec{A}$ has the effect of randomizing $\vec{x}$.
For example, if $\vec{A}$ is i.i.d.\ Gaussian or RRI, then the right singular-vector matrix $\vec{V}$ is Haar distributed, and consequently $\vec{V}\herm\vec{x}$ will be uniformly distributed on the sphere of radius $\|\vec{x}\|$ for any $\vec{x}$,
which means that $\vec{V}\herm\vec{x}$ will behave like an i.i.d.\ Gaussian vector at high dimensions.
In MRI, however, $\vec{A}$ is Fourier structured, and MRI images $\vec{x}$ are also Fourier structured in that they have much more energy near the origin of k-space, with the result that $\vec{V}\herm\vec{x}$ is far from i.i.d.\ Gaussian.
To address this issue, a plug-and-play VAMP approach was proposed in \cite{schniter2017plug} that recovers the \emph{wavelet coefficients} of the image rather than the image itself. 
Importantly, it exploits the fact that---for natural images---the energy of the wavelet coefficients is known to decrease exponentially across the wavelet bands.
By exploiting this additional structure, VAMP can properly handle the joint structure of $\vec{A}$ and $\vec{x}$ that manifests in MRI.  
%Through this approach, the measurement matrix (now a Fourier-Wavelet composition) better randomizes randomizes the signal (now the wavelet coefficients).
In the denoising step, the estimated wavelet coefficients are first transformed to the spatial domain, then denoised, and then transformed back to the wavelet domain.
These transformations are made efficient by the $O(N)$ complexity of the wavelet transform and its inverse.
Additional details are provided in \cite{schniter2017plug}.
\fi % AMP section

%%%%%%%%%%%%%%%%%%%%%%%%%%%%%%%%%%%%%%%%%%%%%%%%%%%
\section{Understanding PnP through Consensus Equilibrium} \label{sec:CE}

The success of the PnP methods in Section~\ref{sec:methods} raises important theoretical questions.
For example, in the case of PnP ADMM, if the plug-in denoiser $\vec{f}$ is not the proximal map of any regularizer $\phi$, then it is not clear what cost function is being minimized (if any) or whether the algorithm will even converge.
Similarly, in the case of RED, if the plug-in denoiser $\vec{f}$ is not the MMSE denoiser $\denoisermmse$, then RED no longer solves the MAP-KDE problem, and it is not clear what RED does solve, or whether a given RED algorithm will even converge.
In this section, we show that many of these questions can be answered through the consensus equilibrium (CE) framework \cite{buzzard2018plug,chan2019performance,sun2019online,reehorst2019regularization}\ocite{wang2018mace}.
We start by discussing CE for the PnP approaches from Section~\ref{sec:PnP} and follow with a discussion of CE for the RED approaches from Section~\ref{sec:RED}.

\subsection{Consensus equilibrium for prox-based PnP} \label{sec:CE PnP}

Let us start by considering the PnP ADMM algorithm \eqref{eq:PnP admm}.
Rather than viewing \eqref{eq:PnP admm} as minimizing some cost function, we can view it as seeking a solution $(\hvec{x}\pnp,\hvec{u}\pnp)$ to 
\begin{subequations} \label{eq:PnP CE}
\begin{eqnarray}
\label{eq:PnP CE loss}
\hvec{x}\pnp &=& \proxloss(\hvec{x}\pnp - \hvec{u}\pnp;\eta)\\
\label{eq:PnP CE reg}
\hvec{x}\pnp &=& \denoiser(\hvec{x}\pnp + \hvec{u}\pnp) ,
\end{eqnarray}
\end{subequations}
which, by inspection, must hold when \eqref{eq:PnP admm} is at a fixed point.
Not surprisingly, by setting $\vec{x}_k=\vec{x}_{k-1}$ in the PnP FISTA algorithm \eqref{eq:PnP fista}, it is straightforward to show that it too seeks a solution to \eqref{eq:PnP CE}.
%A similar procedure can be applied to the primal-dual PnP algorithm from \cite{ono2017primal}.
\textb{It is easy to show that the PnP PDS algorithm \cite{ono2017primal} seeks the same solution.}
With \eqref{eq:PnP CE}, the goal of the prox-based PnP algorithms becomes well defined!
The pair \eqref{eq:PnP CE} reaches a \emph{consensus} in that the denoiser $\denoiser$ and the data fitting operator $\proxloss$ agree on the output $\hvec{x}\pnp$. 
The \emph{equilibrium} comes from the opposing signs on the %``noise-like''
\textb{correction term $\hvec{u}\pnp$: the data-fitting subtracts it while the denoiser adds it.}
%the data fitting operator removes it, while the denoiser adds it back.  

\ifarxiv
\makeblue
Applying the $\vec{h}$ expression from \eqref{eq:proxloss} to \eqref{eq:PnP CE loss}, we find that
\begin{equation}
\hvec{u}\pnp = \frac{\eta}{\sigma^2} \vec{A}\herm(\vec{y}-\vec{A}\hvec{x}\pnp)
\label{eq:PnP CE u} ,
\end{equation}
where $\vec{y}-\vec{A}\hvec{x}\pnp$ is the k-space measurement error and $\vec{A}\herm(\vec{y}-\vec{A}\hvec{x}\pnp)$ is its projection back into the image domain.
%Any components of $\vec{y}-\vec{A}\hvec{x}\pnp$ that are not in the measurement space (i.e., the row-space of $\vec{A}$) will not be present in $\hvec{u}\pnp$.
We now see that $\hvec{u}\pnp$ provides a correction on $\hvec{x}\pnp$ for any components that are inconsistent with the measurements, but it provides no correction for errors in $\hvec{x}\pnp$ that lie outside the row-space of $\vec{A}$.
%Similarly, using \eqref{eq:PnP CE reg}, we can write
%\begin{equation}
%\hvec{u}\pnp &=& (\vec{I}-\denoiser)(\hvec{x}\pnp + \hvec{u}\pnp)
%\label{eq:PnP CE u2} .
%\end{equation}
Plugging \eqref{eq:PnP CE u} back into \eqref{eq:PnP CE reg}, we obtain the image-domain fixed-point equation 
\begin{equation}
\hvec{x}\pnp = \denoiser\left(\left(\vec{I}-\frac{\eta}{\sigma^2}\vec{A}\herm\vec{A}\right)\hvec{x}\pnp + \frac{\eta}{\sigma^2} \vec{A}\herm\vec{y}\right) 
\label{eq:PnP FP1}.
\end{equation}
If $\vec{y}=\vec{Ax}+\vec{w}$, as in \eqref{eq:y}, and we define the PnP error as $\hvec{e}\pnp \defn \hvec{x}\pnp-\vec{x}$, then \eqref{eq:PnP FP1} implies 
\begin{equation}
%\hvec{e}\pnp = \denoiser\left(\vec{x} + \left(\vec{I}-\frac{\eta}{\sigma^2}\vec{A}\herm\vec{A}\right)\hvec{e}\pnp + \frac{\eta}{\sigma^2} \vec{A}\herm\vec{w}\right) - \vec{x}
\hvec{x}\pnp = \denoiser\left(\hvec{x}\pnp +\frac{\eta}{\sigma^2}\vec{A}\herm\left(\vec{w}-\vec{A}\hvec{e}\pnp\right)\right)
\label{eq:PnP FP2},
\end{equation}
%or, equivalently,
%\begin{equation}
%\frac{\eta}{\sigma^2}\vec{A}\herm\left(\vec{A}\hvec{e}\pnp-\vec{w}\right) = (\denoiser-\vec{I})\left(\hvec{x}\pnp -\frac{\eta}{\sigma^2}\vec{A}\herm\left(\vec{A}\hvec{e}\pnp-\vec{w}\right)\right)
%\frac{\eta}{\sigma^2}\vec{A}\herm\vec{A}\hvec{e}\pnp - \frac{\eta}{\sigma^2} \vec{A}\herm\vec{w} = (\denoiser-\vec{I})\left(\vec{x} + \left(\vec{I}-\frac{\eta}{\sigma^2}\vec{A}\herm\vec{A}\right)\hvec{e}\pnp + \frac{\eta}{\sigma^2} \vec{A}\herm\vec{w}\right) 
%\label{eq:PnP FP3},
%\end{equation}
which says that the error $\hvec{e}\pnp$ combines with the k-space measurement noise $\vec{w}$ in such a way that the corresponding image-space correction $\hvec{u}\pnp=\frac{\eta}{\sigma^2}\vec{A}\herm(\vec{w}-\vec{A}\hvec{e}\pnp)$ is canceled by the denoiser $\vec{f}(\cdot)$.
\makeblack
\fi

By viewing the goal of prox-based PnP as solving the equilibrium problem \eqref{eq:PnP CE}, it becomes clear that other solvers beyond ADMM, FISTA, and \textb{PDS} can be used.
For example, it was shown in \cite{buzzard2018plug} that the PnP CE condition \eqref{eq:PnP CE} can be achieved by finding a fixed-point of the system\footnote{The paper \cite{buzzard2018plug} actually considers the consensus equilibrium among $N>1$ agents, whereas here we consider the simple case of $N=2$ agents.} 
\begin{align}
\uvec{z} 
&= (2\mathcal{G}-\vec{I})(2\mathcal{F}-\vec{I})\uvec{z}
\label{eq:PnP CE2}\\
\uvec{z} 
&= \begin{bmatrix}
\vec{z}_1\\
\vec{z}_2
\end{bmatrix}, 
\quad
\mathcal{F}(\uvec{z}) 
= \begin{bmatrix}
\proxloss(\vec{z}_1;\eta)\\
\denoiser(\vec{z}_2)
\end{bmatrix},
\quad
\mbox{and}
\quad
\mathcal{G}(\uvec{z}) 
= \begin{bmatrix}
\frac{1}{2}(\vec{z}_1+\vec{z}_2) \\
\frac{1}{2}(\vec{z}_1+\vec{z}_2) 
\end{bmatrix}.
\end{align}
There exist many algorithms to solve \eqref{eq:PnP CE2}. 
For example, one could use the Mann iteration \ocite{parikh2013proximal}
\begin{align}
\uvec{z}^{(k+1)} = (1-\gamma)\uvec{z}^{k} + \gamma (2\mathcal{G}-\vec{I})(2 \mathcal{F}-\vec{I})\uvec{z}^{(k)},
\quad \text{with $\gamma\in(0,1)$}, 
\end{align}
when $\mathcal{F}$ is nonexpansive.  
The paper \cite{buzzard2018plug} also shows that this fixed point is equivalent to the solution of $\mathcal{F}(\uvec{z}) = \mathcal{G}(\uvec{z})$, in which case Newton's method or other root-finding methods could be applied.  

The CE viewpoint also provides a path to proving the convergence of the PnP ADMM algorithm.
Sreehari et al.\ \cite{sreehari2016plug} used a classical result from convex analysis to show that a sufficient condition for convergence is that i) $\denoiser$ is non-expansive, i.e., $\|\denoiser(\vec{x})-\denoiser(\vec{y})\| \le \|\vec{x}-\vec{y}\|$ for any $\vec{x}$ and $\vec{y}$, and ii) $\denoiser(\vec{x})$ is a sub-gradient of some convex function, i.e., there exists $\varphi$ such that $\denoiser(\vec{x}) \in \partial \varphi(\vec{x})$. 
If these two conditions are met, then PnP ADMM \eqref{eq:PnP admm} will converge to a global solution. 
Similar observations were made in other recent studies, e.g., \cite{chan2019performance}\ocite{teodoro2017convergence}.
That said, Chan et al.\ \cite{chan2017plug} showed that many practical denoisers do not satisfy these conditions, and so they designed a variant of PnP ADMM in which $\eta$ is decreased at every iteration. 
Under appropriate conditions on $\denoiser$ and the rate of decrease, this latter method also guarantees convergence, although not exactly to a fixed point of \eqref{eq:PnP CE} since $\eta$ is no longer fixed.

Similar techniques can be used to prove the convergence of other prox-based PnP algorithms.
For example, under certain technical conditions, including non-expansiveness of $\denoiser$, it was established \cite{sun2019online} that PnP FISTA converges to the same fixed-point as PnP ADMM.

\subsection{Consensus equilibrium for RED} \label{sec:CE RED}
Just as the prox-based PnP algorithms in Section~\ref{sec:PnP} can be viewed as seeking the consensus equilibrium of \eqref{eq:PnP CE}, it was shown in \cite{reehorst2019regularization} that the proximal-gradient and ADMM-based RED algorithms seek the consensus equilibrium $(\hvec{x}\red,\hvec{u}\red)$ of
\begin{subequations} \label{eq:RED CE}
\begin{align}
\hvec{x}\red &= \proxloss(\hvec{x}\red-\hvec{u}\red;\eta) 
\label{eq:RED CE loss}\\
\hvec{x}\red &= \left(\left(1+\frac{1}{L}\right)\vec{I}-\frac{1}{L}\denoiser\right)^{-1}(\hvec{x}\red+\hvec{u}\red) 
\label{eq:RED CE reg}
\end{align}
\end{subequations}
where $\proxloss$ was defined in \eqref{eq:proxloss} and $L$ is the algorithmic parameter that appears in \eqref{eq:red apg}.\footnote{The parameter $L$ also manifests in ADMM RED, as discussed in \cite{reehorst2019regularization}.}
Since \eqref{eq:RED CE} takes the same form as \eqref{eq:PnP CE}, we can directly compare the CE conditions of RED and prox-based PnP.

Perhaps a more intuitive way to compare the CE conditions of RED and prox-based PnP follows from rewriting \eqref{eq:RED CE reg} as  $\hvec{x}\red=\denoiser(\hvec{x}\red) +L\hvec{u}\red$, after which the RED CE condition becomes
\begin{subequations} \label{eq:RED CE3}
\begin{align}
\hvec{x}\red &= \proxloss(\hvec{x}\red-\hvec{u}\red;\eta) 
\label{eq:RED CE loss3}\\
\hvec{x}\red &=\denoiser(\hvec{x}\red) +L\hvec{u}\red 
\label{eq:RED CE reg3} ,
\end{align}
\end{subequations}
which involves no inverse operations.
In the typical case of $L=1$, we see that \eqref{eq:RED CE3} matches \eqref{eq:PnP CE}, except that the \textb{correction} $\hvec{u}\red$ is added \emph{after} denoising in \eqref{eq:RED CE reg3} and \emph{before} denoising in \eqref{eq:PnP CE reg}.

Yet another way to compare the CE conditions of RED and prox-based PnP is to eliminate the $\hvec{u}\red$ variable.
\makeblue
Solving \eqref{eq:RED CE loss3} for $\hvec{u}\red$ gives
\begin{equation}
\hvec{u}\red = \frac{\eta}{\sigma^2} \vec{A}\herm(\vec{y}-\vec{A}\hvec{x}\red)
\label{eq:RED CE u} ,
\end{equation}
which mirrors the expression for $\hvec{u}\pnp$\ifarxiv in \eqref{eq:PnP CE u}\fi.
Then plugging $\hvec{u}\red$ back into \eqref{eq:RED CE reg3} and rearranging, we obtain the fixed-point equation
\begin{equation}
\hvec{x}\red =\denoiser(\hvec{x}\red) + \frac{L\eta}{\sigma^2} \vec{A}\herm(\vec{y}-\vec{A}\hvec{x}\red)
\label{eq:RED CE reg2} ,
\end{equation}
or equivalently
\begin{equation}
\frac{L\eta}{\sigma^2} \vec{A}\herm(\vec{A}\hvec{x}\red-\vec{y})
=\denoiser(\hvec{x}\red) - \hvec{x}\red 
\label{eq:RED CE reg4} ,
\end{equation}
which says that the data-fitting correction (i.e., the left side of \eqref{eq:RED CE reg4}) must balance the  denoiser correction (i.e., the right side of \eqref{eq:RED CE reg4}).
\makeblack

%Solving \eqref{eq:RED CE reg3} for $\hvec{u}\red$ 
%and plugging the result back into \eqref{eq:RED CE loss3} gives the fixed-point condition
%\begin{align}
%\hvec{x}\red 
%&= \proxloss\left( \frac{1}{L}\denoiser(\hvec{x}\red) + \left(1-\frac{1}{L}\right)\hvec{x}\red;\eta\right) \\
%&= \proxloss(\denoiser(\hvec{x}\red);\eta) \text{~~when $L=1$}
%\label{eq:RED CE2} .
%\end{align}
%Applying the same procedure to \eqref{eq:PnP CE} yields the fixed-point condition
%\begin{align}
%\hvec{x}\pnp 
%&= \proxloss\left( (2\vec{I}-\denoiser^{-1})(\hvec{x}\pnp);\eta\right)
%\label{eq:PnP CE3} ,
%\end{align}
%or, equivalently, the fixed-point condition stated earlier in \eqref{eq:PnP CE2}.

The CE framework also facilitates the convergence analysis of RED algorithms.
For example, using the Mann iteration \ocite{parikh2013proximal}, it was proven in \cite{reehorst2019regularization} that when $\denoiser$ is nonexpansive and $L>1$, the PG RED algorithm converges to a fixed point.

\section{Demonstration of PnP in MRI} \label{sec:sim}
\subsection{Parallel cardiac MRI}
\begin{figure}[t]
\centering
\includegraphics[width=5.0in,trim=0 0 0 0mm,clip]{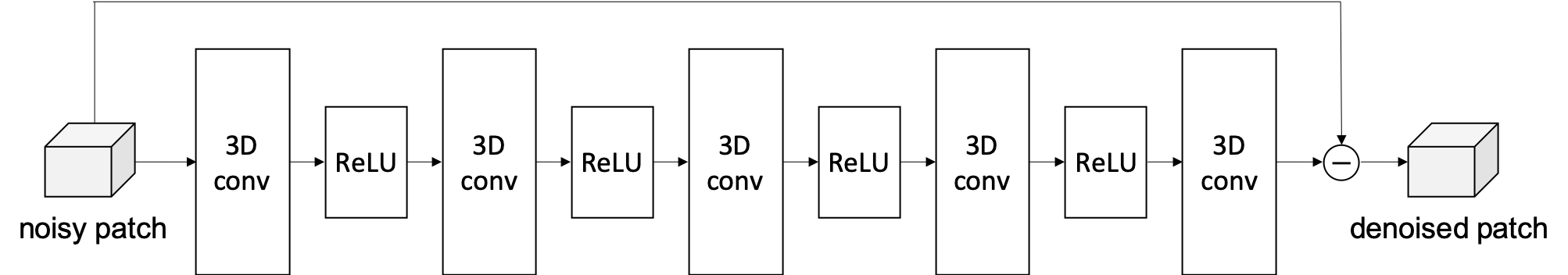} 
\caption{The architecture of the CNN-based cardiac cine denoiser operating on spatiotemporal volumetric patches.}
\label{fig:cnn}
\end{figure}

We now demonstrate the application of PnP methods to parallel cardiac MRI.
%In particular, we demonstrate the recovery of cardiac cine datasets using a learned denoiser.  
Because the signal $\vec{x}$ is a cine (i.e., a video) rather than a still image, there are relatively few options available for sophisticated denoisers. 
Although algorithmic denoisers like BM4D \ocite{maggioni2013nonlocal} have been proposed, they tend to run very slowly, especially relative to the linear operators $\vec{A}$ and $\vec{A}\herm$. 
For this reason, we first trained an application specific CNN denoiser for use in the PnP framework.
The architecture of the CNN denoiser, implemented and trained in PyTorch, is shown in Figure~\ref{fig:cnn}. 

For training, we acquired 50 fully sampled, high-SNR cine datasets from eight healthy volunteers. Thirty three of those were collected on a 3~T scanner\footnote{The 3~T scanner was a Magnetom Prisma Fit from Siemens Healthineers in Erlangen, Germany and the 1.5~T scanner was a Magnetom Avanto from Siemens Healthineers in Erlangen, Germany.} and the remaining 17 were collected on a 1.5~T scanner. Out of the 50 datasets, 28, 7, 7, and 8 were collected in the short-axis, two-chamber, three-chamber, and four-chamber view, respectively. The spatial and temporal resolutions of the images ranged from 1.8~mm to 2.5~mm and from 34~ms to 52~ms, respectively. \textb{The images size ranged from $160\times 130$ to $256 \times 208$ pixels and the number of frames ranged from $15$ to $27$.} For each of the 50 datasets, the reference image series was estimated as the least-squares solution to \eqref{eq:mri}, with the sensitivity maps $\vec{S}_i$ estimated from the time-averaged data using ESPIRiT \ocite{uecker2014espirit}.  
We added zero-mean, complex-valued i.i.d. Gaussian noise to these ``noise-free'' reference images to simulate noisy images with SNR of 24~dB. Using a fixed stride of $30\times 30\times 10$, we decomposed the images into patches of size $55\times 55\times15$. The noise-free and corresponding noisy patches were assigned as output and input to the CNN denoiser, with the real and imaginary parts processed as two separate channels. All 3D convolutions were performed using $3\times 3\times 3$ kernels. There were 64 filters of size $3\times 3\times 3 \times 2$ in the first layer, 64 filters of size $3\times 3\times 3 \times 64$ in the second through fourth layers, and 2 filters of size $3\times 3\times 3 \times 64$ in the last layer. We set the minibatch size to four and used the Adam optimizer \ocite{kingma2015adam} with a learning rate of $1\times 10^{-4}$ over 400 epochs. The training process was completed in 12 hours on a workstation equipped with a single NVIDIA GPU (GeForce RTX 2080 Ti). 

\begin{figure}[t]
\centering
\includegraphics[width=2.5in,trim=0 0 0 0mm,clip]{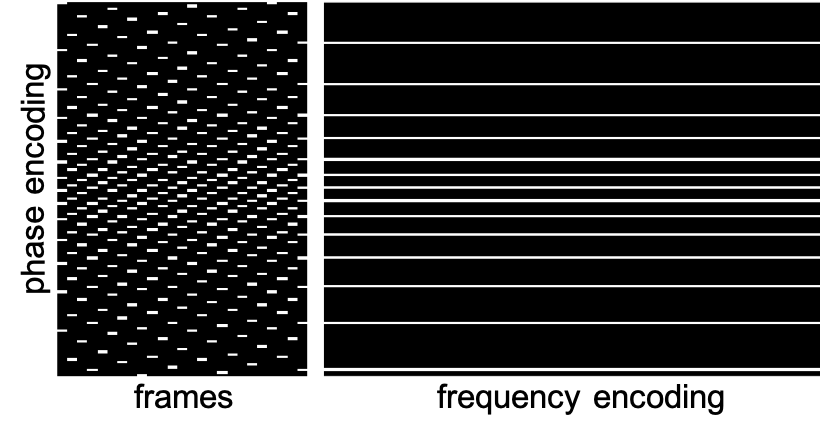} 
\caption{Two different views of the 3D sampling pattern used to retrospectively undersample one of the four test datasets at $R=10$. The undersampling was performed only in the phase encoding direction and the pattern was varied across frames. In this example, the number of frequency encoding steps, phase encoding steps, and frames are 192, 144, and 25, respectively.}
\label{fig:sampling}
\end{figure}

For testing, we acquired four fully sampled cine datasets from two different healthy volunteers, with two image series in the short-axis view, one image series in the two-chamber view, and one image series in the four-chamber view. The spatial and temporal resolutions of the images ranged from 1.9~mm to 2~mm and from 37~ms to 45~ms, respectively. For the four datasets, the space-time signal vector, $\vec{x}$, in \eqref{eq:y} had dimensions of $192\times 144\times 25$, $192\times 144\times 25$, $192\times 166\times 16$, and $192\times 166\times 16$, respectively, with the last dimension representing the number of frames. The datasets were retrospectively downsampled at acceleration rates, $R$, of 6, 8, and 10 using pseudo-random sampling \cite{ahmad2015variable}. A representative sampling pattern used to undersample one of the datasets is shown in Figure~\ref{fig:sampling}. The data were compressed to $C=12$ virtual coils for faster computation \ocite{buehrer2007array}. The measurements were modeled as described in~\eqref{eq:mri}, with the sensitivity maps, $\vec{S}_i$, estimated from the time-averaged data using ESPIRiT. 

For compressive MRI recovery, we used PnP ADMM from~\eqref{eq:PnP admm} with $\vec{f}(\cdot)$ as the CNN-based denoiser described above; we will refer to the combination as PnP-CNN. We employed a total of 100 ADMM iterations, and in each ADMM iteration, we performed four steps of CG to approximate~\eqref{eq:proxloss}, for which we used $\sigma^2=1=\eta$. 
\ifarxiv(See Figure~\ref{fig:nmse_nu} for the effect of $\sigma^2/\eta$ \textb{on the final NMSE and the convergence rate}.)\fi
We compared this PnP method to three CS-based methods: \ifarxiv{\textb{CS with an undecimated wavelet transform (CS-UWT)}}\else{CS-UWT}\fi, \ifarxiv{\textb{CS with total variation (TV)}}\else{CS-TV}\fi,\footnote{\textb{Note that sometimes UWT and TV are combined \cite{lustig2007sparse}.}} and a low-rank plus sparse (L+S) method (see, e.g., \cite{otazo2015lps}). 
We also compared to PnP-UWT and the transform-learning \ifarxiv{\cite{wen2019transform}}\else{(see the overview by Wen, Ravishankar, Pfister, and Bresler in this special issue) }\fi method LASSI \cite{ravishankar2017lowrank}. 

For PnP-UWT, we used PnP FISTA from~\eqref{eq:PnP fista} with $\vec{f}(\cdot)$ implemented as $\denoisertdt$ given in~\eqref{eq:denoisertdt}, i.e., bFISTA. 
A three-dimensional single-level Haar UWT was used as $\vec{\Psi}$ in~\eqref{eq:denoisertdt}.
For CS-TV, we used a 3D finite-difference operator for $\vec{\Psi}$ in the regularizer $\phi(\vec{x})=\|\vec{\Psi x}\|_1$, and for CS-UWT, we used the aforementioned UWT instead. For both CS-TV and CS-UWT, we used monotone FISTA \cite{tan2014fista} to solve the resulting convex optimization problem \eqref{eq:reg_inverse}. 
For L+S, the method by Otazo et al.~\cite{otazo2015lps} was used. 
The regularization weights for CS-UWT, PnP-UWT, CS-TV, and L+S were manually tuned to %minimize the normalized MSE (NMSE) 
maximize the \textb{reconstruction SNR (rSNR)}\footnote{\textb{rSNR is defined as $\|\vec{x}\|^2/\|\hvec{x}-\vec{x}\|^2$, where $\vec{x}$ is the true image and $\hvec{x}$ is the estimate.}}
for \textb{Dataset \#3} at $R=10$. 
For LASSI we used the authors' implementation at \url{https://gitlab.com/ravsa19/lassi}, and we did our best to manually tune all available parameters.

\begin{figure}[h!]
\centering
\includegraphics[width=6.0in,trim=0 0 0 0mm,clip]{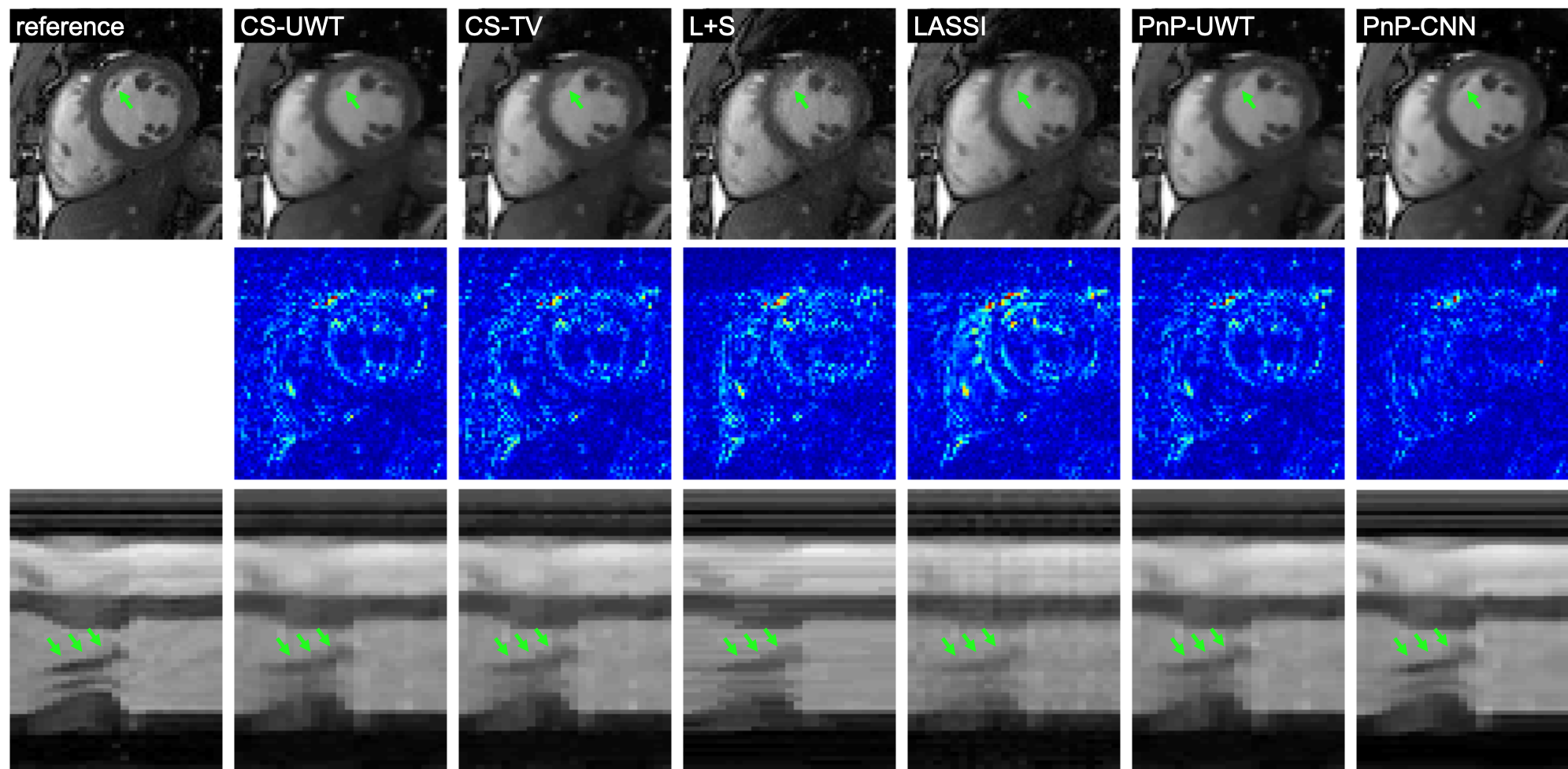} 
\caption{Results from cardiac cine Dataset \#1 at $R=10$. Top row: a representative frame from the fully sampled reference and various recovery methods. The green arrow points to an image feature that is preserved only by PnP-CNN and not by other methods. Middle row: error map $\times 6$. Bottom row: temporal frame showing the line drawn horizontally through the middle of the image in the top row, with the time dimension along the horizontal axis. The arrows point to the movement of the papillary muscles, which are more well-defined in PnP-CNN.}
\label{fig:pnp_example}
\end{figure}

\begin{figure}[h!]
\centering
\includegraphics[width=3.0in,trim=0 0 0 0mm,clip]{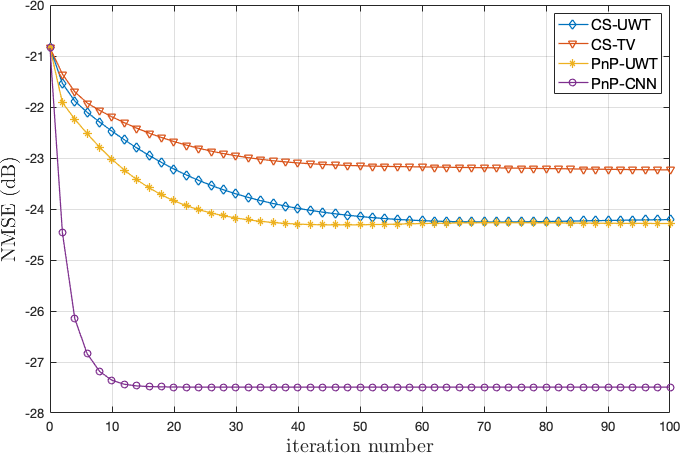} 
\caption{NMSE versus iteration for two PnP and two CS algorithms on the cardiac cine recovery Dataset \#3 at $R=10$.}
\label{fig:nmse_iter}
\end{figure}

The \textb{rSNR} values are summarized in Table \ref{tab:nmse}. For all four datasets and three acceleration rates, PnP-CNN exhibited the %lowest NMSE 
\textb{highest rSNR}
with respect to the fully sampled reference. 
Also, compared to the CS methods and PnP-UWT, which uses a more traditional denoiser based on soft-thresholding of UWT coefficients, PnP-CNN was better at preserving anatomical details of the heart; see Figure~\ref{fig:pnp_example}. 
The performance of PnP-UWT was similar to that of CS-UWT. 
Figure \ref{fig:nmse_iter} plots NMSE as a function of the number of iterations for the CS and PnP methods. 
Since the CS methods were implemented using CPU computation and the PnP methods were implemented using GPU computation, a direct runtime comparison was not possible. 
We did, however, compare the per-iteration runtime of PnP ADMM for two different denoisers: the CNN and UWT-based $\denoisertdt$ described earlier in this section. 
When the CNN denoiser was replaced with the UWT-based $\denoisertdt$, the per-iteration runtime changed from 2.05~s to 2.1~s, implying that the two approaches have very similar computational costs. 
\ifarxiv

For PnP-CNN, Figure~\ref{fig:nmse_nu} shows the dependence of the final NMSE \textb{($=\text{rSNR}^{-1}$)} \textb{and of the convergence rate} on $\sigma^2/\eta$ for one of the testing datasets included in this study. 
Overall, final NMSE varies less than 0.5~dB for $\sigma^2/\eta \in [0.5, 2]$ for all four datasets and all three acceleration rates.
Figure~\ref{fig:cg_gd_inner} compares CG and GD when solving \eqref{eq:proxloss}. 
To this end, NMSE vs.\ runtime is plotted for different numbers of CG or GD inner-iterations for Dataset~\#3 at $R=10$. 
For GD, the step-size was manually optimized. 
Figure~\ref{fig:cg_gd_inner} suggests that it is best to use a $1$ to $4$ inner iterations of either GD or CG; using more inner iterations slows the convergence rate without improving the final performance.

\else
The extended version of this paper \cite{ahmad2019plug} shows the results of experiments that investigate the effect of $\sigma^2/\eta$ on the final NMSE and the convergence rate.
Overall, final NMSE varies less than 0.5~dB for $\sigma^2/\eta \in [0.5, 2]$ for all four datasets and all three acceleration rates, and the convergence rate is nearly the same.
The extended version also explores the use of CG versus GD when solving \eqref{eq:proxloss} in PnP ADMM.
The results suggest that $1$ to $4$ inner iterations of either method are optimal; more inner iterations slows the overall convergence time.
\fi
The results in this section, although preliminary, highlight the potential of PnP methods for MRI recovery of cardiac cines. 
By optimizing the denoiser architecture, the performance of PnP-CNN may be further improved. 
%Finally, additional studies are needed to further evaluate the performance of PnP methods compared to other deep learning techniques for MRI reconstruction.

\begin{table}[h!]
\centering
\adjustbox{width=4in,keepaspectratio}{
\begin{tabular}{ |c||c|c|c|c|c|c| }
 \hline
 Acceleration & CS-UWT & CS-TV & L+S & LASSI & PnP-UWT & PnP-CNN\\
 \hline
 \hline
 \multicolumn{7}{|c|}{Dataset \#1 (short-axis)} \\ %FID17464
 \hline
 $R=6$    & 30.10 & 29.03 & 30.97 & 27.09 & 30.18 & \textbf{31.82}\\
 $R=8$    & 28.50 & 27.35 & 29.65 & 25.91 & 28.60 & \textbf{31.25}\\
 $R=10$   & 26.94 & 25.78 & 28.29 & 24.98 & 27.06 & \textbf{30.46}\\
 \hline
 \multicolumn{7}{|c|}{Dataset \#2 (short-axis)} \\ %FID17462
 \hline
 $R=6$    & 29.23 & 28.27 & 29.73 & 25.87 & 29.29 & \textbf{30.81}\\
 $R=8$    & 27.67 & 26.65 & 28.23 & 24.54 & 27.75 & \textbf{30.17}\\
 $R=10$   & 26.12 & 25.11 & 26.89 & 23.61 & 26.22 & \textbf{29.21}\\
 \hline
 \multicolumn{7}{|c|}{Dataset \#3 (two-chamber)} \\ %FID09590
 \hline
 $R=6$    & 27.33 & 26.38 & 27.83 & 24.97 & 27.38 & \textbf{29.36}\\
 $R=8$    & 25.63 & 24.63 & 26.30 & 23.52 & 25.69 & \textbf{28.50}\\
 $R=10$   & 24.22 & 23.24 & 24.93 & 22.51 & 24.28 & \textbf{27.49}\\
 \hline
 \multicolumn{7}{|c|}{Dataset \#4 (four-chamber)} \\ %FID09578
 \hline
 $R=6$    & 30.41 & \textb{29.63} & 30.62 & 27.62 & 30.60 & \textbf{32.19}\\
 $R=8$    & 28.68 & 27.76 & 29.00 & 26.33 & 28.94 & \textbf{31.42}\\
 $R=10$   & 27.09 & 26.18 & 27.60 & 25.24 & 27.37 & \textbf{30.01}\\
 \hline
\end{tabular}}
\smallskip
\caption{\textb{rSNR} (dB) of MRI cardiac cine recovery from four test datasets.}
\label{tab:nmse}
\end{table}

\ifarxiv
%----------------------------------------------------
\begin{figure}[h!]
\centering
\includegraphics[width=5in,trim=0 0 0 0mm,clip]{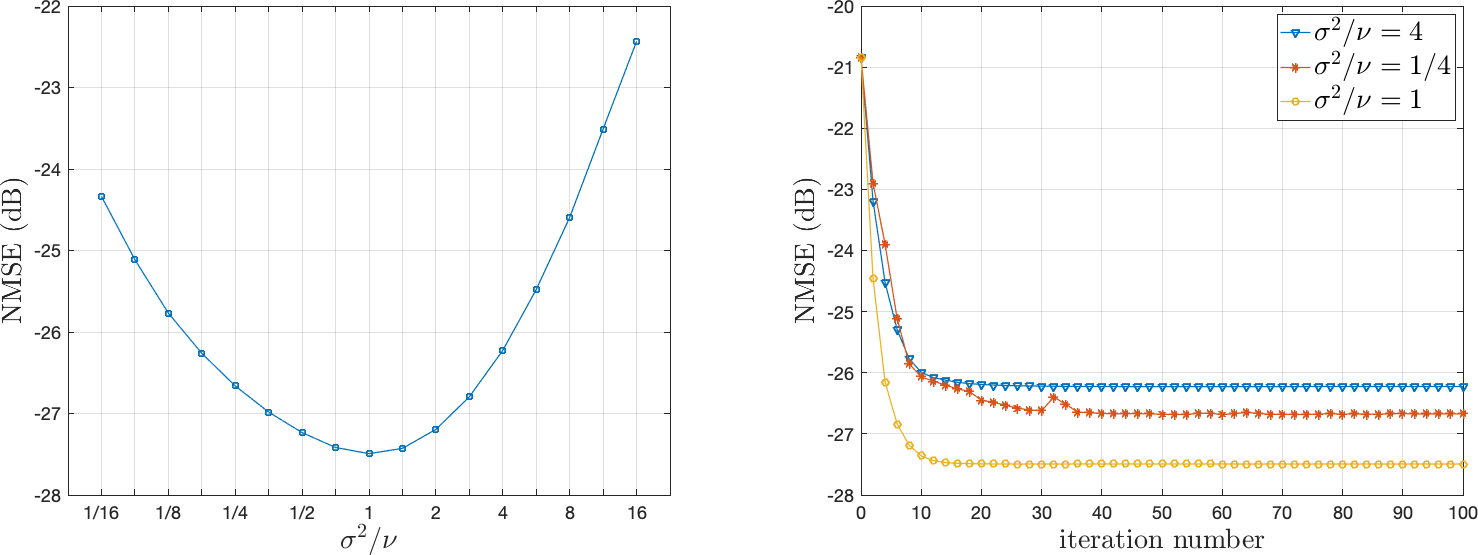} 
\caption{For PnP-CNN recovery of cardiac cine Dataset \#3 at $R=10$, the change in the final NMSE (after 100 iterations) as a function of $\sigma^2/\eta$ (left) \textb{and the NMSE versus iteration for several $\sigma^2/\eta$ (right).}}
\label{fig:nmse_nu}
\end{figure} 

\begin{figure}[h!]
\centering
\includegraphics[width=4.5in,trim=0 0 0 0mm,clip]{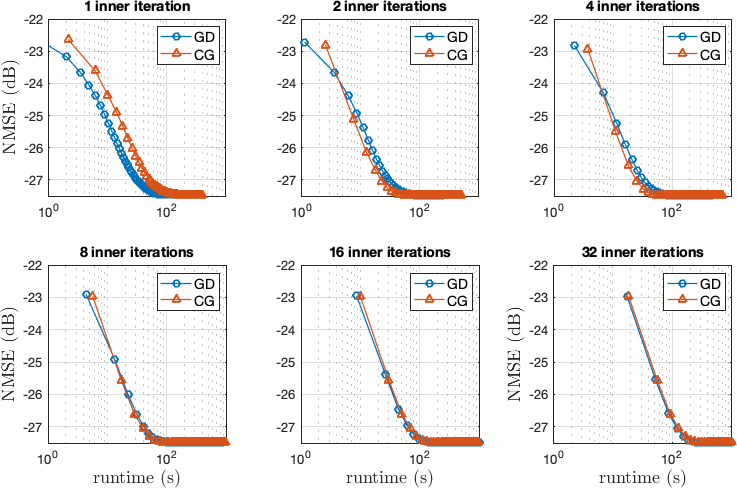} 
\caption{NMSE versus time PnP ADMM with different numbers of CG and GD inner iterations on cardiac cine Dataset \#3 at $R=10$.}
\label{fig:cg_gd_inner}
\end{figure}
%----------------------------------------------------
\fi 

%=================================================================
\makeblue
\subsection{Single-coil fastMRI knee data}
In this section, we investigate recovery of 2D knee images from the single-coil fastMRI dataset \cite{zbontar2018fastmri}.
This dataset contains fully-sampled k-space data that are partitioned into 34\,742 training slices and 7\,135 testing slices.
The Cartesian sampling patterns from \cite{zbontar2018fastmri} were used to achieve acceleration rate $R=4$.

We evaluated PnP using the ADMM algorithm with a learned DnCNN \cite{zhang2017beyond} denoiser. 
To accommodate complex-valued images, DnCNN was configured with two input and two output channels.
The denoiser was then trained using only the central slices of the 3~T scans without fat-suppression from the training set, comprising a total of 267 slices (i.e., $<1\%$ of the total training data).
The training-noise variance and the PnP ADMM tuning parameter $\sigma^2/\eta$ were manually adjusted in an attempt to maximize rSNR.

PnP was then compared to the TV and U-Net baseline methods described and configured in \cite{zbontar2018fastmri}.
%For TV, the tuning parameter was selected as in \cite{zbontar2018fastmri} for images without fat-suppression (i.e., $\lambda = 10^{-2}$).
%The baseline U-net takes the magnitude of the zero-filled image as the input and attempts to reproduce the original at the output.
%The network parameters are adjusted using the $\ell_1$ loss between the fully-sampled image and the reconstruction.
For example, 128 channels were used for the U-Net's first layer, as recommended in \cite{zbontar2018fastmri}.
%Further details of the U-Net are available in \cite{zbontar2018fastmri}.
We then trained three versions of the U-Net.
The first version was trained on the full fastMRI training set\footnote{The full fastMRI training set includes 1.5~T and 3~T scans, with and without fat suppression, and an average of 36 slices per volume.} with random sampling masks.
The second U-Net was trained on the full fastMRI training set, but with a fixed sampling mask.
The third U-Net was trained using only the central slices of the 3~T scans without fat-suppression (i.e., the same data used to train the DnCNN denoiser) and with a fixed sampling mask.

To evaluate performance, we used the central slices of the non-fat-suppressed 3~T scans from the validation set, comprising a total of 49 slices.
The evaluation considered both random sampling masks and the same fixed mask used for training.
The resulting average rSNR and SSIM scores are summarized in Table~\ref{tab:fastMRI}.
The table shows that PnP-CNN performed similarly to the U-Nets and significantly better than TV.
In particular, PnP-CNN achieved the highest rSNR score with both random and fixed testing masks, and the U-Net gave slightly higher SSIM scores in both tests.
Among the U-Nets, the version trained with a fixed sampling mask and full data gave the best rSNR and SSIM performance when testing with the same mask,
but its performance dropped considerable when testing with random masks.
Meanwhile, the U-Net trained with the smaller data performed significantly worse than the other U-Nets, with either fixed or random testing masks.
And although this latter U-Net used exactly the same training data as the PnP-CNN method, it was not competitive with PnP-CNN.
Although preliminary, these results suggest that i) PnP methods are much less sensitive to deviations in the forward model between training and testing, and that ii) PnP methods are effective with relatively small training datasets.

\begin{table}
\makeblue
\centering
\adjustbox{width=5in,keepaspectratio}{
\begin{tabular}{|l||c|c||c|c|}
\hline
& \multicolumn{2}{|c||}{Random testing masks} & \multicolumn{2}{|c|}{Fixed testing mask}\\ \cline{2-5}
& rSNR (dB) & SSIM & rSNR (dB) & SSIM\\ \hline\hline
CS-TV & 17.56 & 0.647 & 18.16 & 0.654\\
U-Net: Random training masks, full training data & 20.76 & \bf{0.772} & 20.72 & 0.768\\
U-Net: Fixed training mask, full training data & 19.63 & 0.756 & 20.82 & \bf{0.770}\\
U-Net: Fixed training mask, smaller training data & 18.90 & 0.732 & 19.67 & 0.742 \\ 
PnP-CNN & \bf{21.16} & 0.758 & \bf{21.14} & 0.754\\
\hline
\end{tabular}}
\smallskip
\caption{rSNR and SSIM for fastMRI single-coil test data with $R=4$. 
%Results are from central slices of the validation set for 3~T scans without fat-suppression.
}
\label{tab:fastMRI}
\end{table}

\makeblack
%=================================================================

\section{Conclusion}

PnP methods present an attractive avenue for compressive MRI recovery. 
In contrast to traditional CS methods, PnP methods can exploit richer image structure by using state-of-the-art denoisers. 
To demonstrate the potential of such methods for MRI reconstruction, we used PnP to recover cardiac \textb{cines and knee images} from highly undersampled datasets. 
With application-specific CNN-based denoisers, PnP was able to significantly outperform traditional CS methods \textb{and to perform on par with modern deep-learning methods, but with considerably less training data}. 
%With many implementation choices and demonstrable convergence analysis from the equilibrium perspective,
The time is ripe to investigate the potential of PnP methods for a variety of MRI applications.

%=====================================================================
\singlespacing
\footnotesize
%\clearpage
\makeblack
\bibliographystyle{IEEEtran}
\bibliography{bibl.bib}

\end{document}